\PassOptionsToPackage{unicode}{hyperref}
\PassOptionsToPackage{hyphens}{url}
\PassOptionsToPackage{dvipsnames,svgnames,x11names}{xcolor}
\documentclass[
  11pt,
]{article}
\usepackage{xcolor}
\usepackage[margin=1in]{geometry}
\usepackage{amsmath,amssymb}
\usepackage{iftex}
\ifPDFTeX
  \usepackage[T1]{fontenc}
  \usepackage[utf8]{inputenc}
  \usepackage{textcomp} 
\else 
  \usepackage{unicode-math} 
  \defaultfontfeatures{Scale=MatchLowercase}
  \defaultfontfeatures[\rmfamily]{Ligatures=TeX,Scale=1}
\fi
\usepackage{lmodern}
\ifPDFTeX\else
\fi
\IfFileExists{upquote.sty}{\usepackage{upquote}}{}
\IfFileExists{microtype.sty}{
  \usepackage[]{microtype}
  \UseMicrotypeSet[protrusion]{basicmath} 
}{}
\makeatletter
\@ifundefined{KOMAClassName}{
  \IfFileExists{parskip.sty}{%
    \usepackage{parskip}
  }{
    \setlength{\parindent}{0pt}
    \setlength{\parskip}{6pt plus 2pt minus 1pt}}
}{
  \KOMAoptions{parskip=half}}
\makeatother
\usepackage{longtable,booktabs,array}
\usepackage{caption}
\usepackage{calc} 
\usepackage{etoolbox}
\makeatletter
\patchcmd\longtable{\par}{\if@noskipsec\mbox{}\fi\par}{}{}
\makeatother
\IfFileExists{footnotehyper.sty}{\usepackage{footnotehyper}}{\usepackage{footnote}}
\makesavenoteenv{longtable}
\usepackage{graphicx}
\makeatletter
\newsavebox\pandoc@box
\newcommand*\pandocbounded[1]{
  \sbox\pandoc@box{#1}%
  \Gscale@div\@tempa{\textheight}{\dimexpr\ht\pandoc@box+\dp\pandoc@box\relax}%
  \Gscale@div\@tempb{\linewidth}{\wd\pandoc@box}%
  \ifdim\@tempb\p@<\@tempa\p@\let\@tempa\@tempb\fi
  \ifdim\@tempa\p@<\p@\scalebox{\@tempa}{\usebox\pandoc@box}%
  \else\usebox{\pandoc@box}%
  \fi%
}
\def\fps@figure{htbp}
\makeatother
\providecommand{\tightlist}{%
  \setlength{\itemsep}{0pt}\setlength{\parskip}{0pt}}
\usepackage[numbers]{natbib}
\usepackage{bookmark}
\IfFileExists{xurl.sty}{\usepackage{xurl}}{} 
\hypersetup{
  pdftitle={The Horizon Gap: Planning{,} Memory{,} Execution{,} Training{,} and Evaluation for Long-Horizon LLM Agents},
  pdfauthor={Mingguang Chen --- DeepGrounding (corresponding: deepgroundingai@gmail.com); Licheng Wang --- AlphaAvatar; Bo Qu --- DeepGrounding},
  colorlinks=true,
  linkcolor={Maroon},
  filecolor={Maroon},
  citecolor={Blue},
  urlcolor={Blue},
  pdfcreator={LaTeX via pandoc}}

\title{The Horizon Gap: Planning, Memory, Execution, Training, and
Evaluation for Long-Horizon LLM Agents}
\author{Mingguang Chen\textsuperscript{1,$*$} \quad Licheng Wang\textsuperscript{2} \quad Bo Qu\textsuperscript{1}\\[6pt]
{\small \textsuperscript{1}DeepGrounding \quad \textsuperscript{2}AlphaAvatar}\\[2pt]
{\small \textsuperscript{$*$}Corresponding author. Email: \href{mailto:deepgroundingai@gmail.com}{deepgroundingai@gmail.com}}}
\date{July 2026}

\begin{document}
\maketitle

\begin{abstract}
Frontier language models solve, in a single forward pass, reasoning
problems that would have been research contributions a few years ago ---
yet the same models, embedded in an agent loop and asked to complete a
task spanning hours rather than seconds, fail in ways no single-step
benchmark reveals: losing track of an earlier decision, declaring a
half-finished job done, or quietly drifting from the goal they were
given. We call the distance between single-step capability and reliable
long-task completion the \textbf{horizon gap}, and survey \textbf{1,547
arXiv papers (2024-2026)} --- collected via a systematic eight-thread
seed harvest with a disclosed, quantified two-stage bleed filter (26.8\%
of raw hits excluded as off-topic), extended by a targeted supplement
into a badly under-covered theory/safety category --- that map the
field's response to it. We first disambiguate three properties the
literature routinely conflates: \emph{long-horizon} (a property of the
task, its required number of steps), \emph{long-context} (a property of
the model, how many tokens it can attend to at once), and
\emph{long-term memory} (a property of the system, whether information
persists across steps or sessions) --- logically independent axes that a
single ``long-horizon'' label obscures. We then organize the corpus into
six categories that track a long-horizon task's lifecycle: planning and
decomposition, memory and context management, execution control and
recovery, training for long horizons, evaluation and measurement, and
the foundations, limits, and safety of running agents unattended for
extended periods --- crossed with a second axis, \emph{where the horizon
is carried} (within one context, within one task via a harness that
exceeds the context window, or persistently across tasks and sessions),
that organizes every technical section. Across all six, we find the same
structural pattern: outcome-only signals --- a single reward, a single
pass/fail check --- grow uninformative as horizon grows, and the field's
response, whether in training (process reward models and credit
assignment) or evaluation (trajectory-level diagnostics that supersede
pass/fail benchmarks), is to manufacture denser, step-level signal in
its place. We treat the critical and diagnostic literature ---
demonstrations that self-correction, benchmark scores, or training
signals do not mean what they are assumed to mean --- as a first-class
thread throughout rather than a separate critique chapter, since a
survey that segregated critique from method in this field would
routinely split single papers across two chapters. We close by naming
what we take to be the field's most consequential open measurement
problems: how much long-horizon capability lives in the underlying model
versus the harness wrapped around it, the risk of correlated measurement
bias where the process-level signals used to train long-horizon agents
and those used to evaluate them rest on shared assumptions about what
counts as progress, and whether long-horizon reliability admits any
general predictive theory at all.
\end{abstract}

\section{1. Introduction}\label{introduction}

A frontier language model can now solve, in a single forward pass,
reasoning problems that would have been research contributions a few
years ago. The same model, dropped into an agent loop and asked to keep
a multi-hour software-engineering task on track, still fails in ways a
junior engineer would not --- losing track of an earlier decision,
declaring victory on a half-finished patch, or quietly drifting from the
goal it was given. Measurement work tracking the length of task (in
human-time-to-complete) that frontier models can complete at a fixed
reliability threshold reports this length growing at a measured, roughly
exponential rate across recent model generations \citep{kwa2025metr} ---
a striking trend we treat, per this survey's own language calibration
(§2.2), as \emph{measured and reported}, not as a settled law, since
what counts as ``the same'' task across model eras and whether
human-time-to-complete on a benchmark generalizes to messier real
deployments are both contested. But even taking the trend at face value,
it describes model capability outpacing something else: the reliable
\emph{completion} of long tasks in deployment, where harness failures,
evaluation gaps, and accumulating error still bite in ways single-step
benchmarks never see. We call this the \textbf{horizon gap} --- the
distance between what a model can do in one step and what a system built
around it can reliably finish over many --- and argue it is the primary
bottleneck standing between current agent capability and durable
real-world deployment.

This survey maps the research literature responding to that gap. We
assembled a corpus of \textbf{1,547 arXiv papers (2024-2026)} --- a
systematic seed harvest across eight threads of the long-horizon-agent
literature (1,419 papers, after a two-stage bleed filter removed 26.8\%
of raw hits judged off-topic, primarily classical multi-agent-RL,
robotic manipulation, and time-series-forecasting work this survey
explicitly scopes out), extended by a targeted supplemental harvest of
128 papers aimed at directions the reclassification exposed as
under-covered, chiefly the theory and safety literature of §8 --- and
classified all of them into \textbf{six technical categories} (Table 1,
Figure 1). Three observations motivate the paper's structure.

\textbf{First, the field's vocabulary conflates three logically
independent properties.} ``Long- horizon,'' ``long-context,'' and
``long-term memory'' are used almost interchangeably in casual
discussion, but a task's horizon (how many steps it requires), a model's
context length (how many tokens it can attend to at once), and a
system's memory (whether information persists across steps or sessions)
can vary independently --- a harness can carry a long-horizon task
through a short context via aggressive summarization, and a system can
have a huge context window and no persistent memory at all. §2 fixes
this and several other overloaded terms (agent vs.~harness, episode
vs.~session, coherence, autonomy time) before using them, because
imprecision here is not merely stylistic: it is the reason a ``memory''
paper and a ``long-context'' paper are so often cited as if they
addressed the same problem when they do not.

\textbf{Second, no single component is where long-horizon tasks succeed
or fail --- the whole pipeline is.} We organize the survey around six
categories that track a task's lifecycle: \emph{planning} decides what
to do (§3); \emph{memory} decides what information that decision draws
on (§4); \emph{execution control} decides how the resulting actions are
run and recovered from when they fail (§5); \emph{training} decides how
the underlying policy learns to act well over many steps (§6);
\emph{evaluation} decides whether any of this actually works (§7); and
\emph{foundations} covers the theory of why long horizons degrade
performance at all, and the oversight problem of running an agent for a
long time without a human watching every step (§8). Existing surveys
cover individual pieces of this pipeline --- planning strategies, memory
mechanisms, GUI-agent methods, agentic reinforcement learning --- in
isolation (§2.4); this survey's contribution is to organize the full
pipeline around one cross-cutting question asked of every stage:
\emph{what breaks first as the required horizon grows, and what
compensates?}

\textbf{Third, the critical and diagnostic literature is not a footnote
--- it is often the most important literature in each category.} A
recurring pattern across §§3-8 is that the most consequential recent
papers are not new methods but rigorous demonstrations that an earlier
method, benchmark, or metric does not do what it claims: intrinsic
self-correction does not reliably improve reasoning without external
grounding \citep{huang2024cannotselfcorrect}; a substantial share of
reported SWE-bench solves reflect test-suite weakness or training-data
leakage rather than genuine issue resolution
\citep{wang2025solved, liang2025swe}; and agentic benchmarks broadly
have systematic construction problems that best practices are only now
starting to name explicitly \citep{zhu2025establishing}. We treat this
diagnostic thread as a first-class citizen throughout rather than
segregating it into a single critique section, because in this
literature the critique of a method and the state of the art in that
method's category are often the same papers.

\textbf{Contributions.} (1) A disambiguation of long-horizon,
long-context, and long-term memory, and a six-category taxonomy ---
planning, memory, execution, training, evaluation, foundations ---
crossed with a second axis (where the horizon is carried:
within-context, within-task-beyond- context, or cross-task-persistent)
that organizes every technical section (§2). (2) A systematic map of
1,547 papers into that taxonomy, with a disclosed, quantified
query-bleed rate and an explicit two-stage construction (§2.2, Figure 2,
Figure 6). (3) A synthesis identifying two specific unresolved
measurement problems --- how much of long-horizon capability lives in
the model versus the harness (§5, §9), and the risk of correlated
measurement bias between the process-level signals used to train
long-horizon agents and those used to evaluate them (§6, §7, §9) ---
together with a cross-cutting observation that the execution
\emph{trajectory}, not the final outcome, is becoming the field's shared
unit of analysis (§9). (4) An explicit positioning of this survey
against the recursive-self-improvement literature it borders but does
not overlap with, self-cited and scoped in §2.4 and §6.

\section{2. Preliminaries}\label{preliminaries}

\subsection{2.1 Definitions}\label{definitions}

The literature on long-horizon agents uses a small set of terms so
loosely that surveys risk adding noise rather than removing it. We fix
working definitions here and use them consistently for the rest of the
paper.

A \textbf{task} is a specification of a desired end state together with
a checkable completion criterion --- ``resolve this GitHub issue,''
``book an itinerary satisfying these constraints,'' ``reduce this eval's
error rate.'' A task's \textbf{horizon} is the number of sequential
decisions, actions, or environment interactions required to go from the
initial state to task completion; it is a property of the task (and the
environment it is embedded in), not of any particular system attempting
it. An \textbf{episode} is one complete attempt at a task, from a
defined start state to a terminal state --- success, failure, or
time-out --- the unit of account inherited from reinforcement learning.
A \textbf{session} is a continuous interactive period between a user (or
an orchestrating process) and an agent, bounded by the interface layer
rather than by the task: a session may contain zero, one, or many
episodes, and a single episode may span multiple sessions if the
system's state persists across the boundary. Conflating episode and
session is a common source of confusion in the memory literature (§4),
where ``long-term'' sometimes means ``beyond this episode'' and
sometimes means ``beyond this session.''

We also distinguish an \textbf{agent} --- the policy that decides which
action to take next, i.e. the model together with whatever
prompting/decoding strategy turns its outputs into actions --- from its
\textbf{harness} (also called scaffold): the surrounding software that
converts a single-forward-pass model into a system capable of taking
many actions over time --- the control loop, tool-execution sandbox,
memory read/write layer, retry and recovery logic, and any sub-agent
orchestration. This distinction matters because a recurring finding
across §§3--5 is that harness engineering, not the underlying model, is
often the binding constraint on how long a task an agent can complete
--- the same model dropped into a better harness completes measurably
longer tasks.

The paper's central disambiguation is among three axes that the field's
vocabulary routinely conflates:

\begin{itemize}
\tightlist
\item
  \textbf{Long-context} is a property of the model/serving stack: how
  many tokens a single forward pass can attend to.
\item
  \textbf{Long-horizon} is a property of the \emph{task}: how many
  sequential decisions or actions it requires, independent of whether
  the resulting interaction trace fits inside one context window.
\item
  \textbf{Long-term memory} is a property of the \emph{system}: whether
  information available at step or session \(t\) remains available (in
  some form --- verbatim in context, retrieved from an external store,
  or baked into weights) at step or session \(t+k\).
\end{itemize}

These are logically independent. A task can be long-horizon but
short-context, if a harness aggressively summarizes or discards
intermediate state between steps (at some risk to correctness, §4). A
system can have a very long context window and still have no long-term
memory in any interesting sense, if every session starts from a blank
slate. And a system can exhibit long-term memory without ever facing a
long-horizon task, if it merely caches user preferences across
otherwise-independent short interactions. Table 1 and Figure 1 organize
the survey along a related but distinct axis --- where the extra horizon
is \emph{carried} --- which we introduce in §2.3.

Two further terms recur across sections. \textbf{Coherence} denotes the
property that an agent's actions across steps remain consistent with a
single evolving representation of the goal and the plan toward it,
rather than silently drifting, forgetting earlier commitments, or
contradicting itself; loss of coherence is the qualitative failure mode
that long-horizon degradation curves are trying to measure
quantitatively (§8). \textbf{Autonomy time} (or \emph{time horizon}, in
the sense popularized by measurement work discussed in §7 and §8) is an
empirical capability metric: the length of task, measured in
human-time-to-complete, that an agent can complete at a fixed
reliability threshold (e.g., 50\%) --- a way of putting a number on
``how long a horizon can this system actually carry,'' as opposed to the
task-side horizon defined above.

\subsection{2.2 Corpus construction}\label{corpus-construction}

The corpus was built in two stages: a systematic seed harvest, then a
targeted supplemental harvest aimed at categories the first stage
under-covered. Both stages, and the classification step between them,
were carried out by a single annotator (the author); all classification
rules are released as scripts alongside the corpus (§ Data
Availability), and the noise this process introduces is quantified below
rather than assumed away.

\textbf{Seed harvest.} We queried the arXiv API across eight threads
covering the literature's recognized sub-directions --- long-horizon
tasks and agents; LLM agent planning and task decomposition; agent
memory and context management; agentic reinforcement learning and credit
assignment; agent benchmarks and evaluation; multi-agent orchestration
and scaffolding; error accumulation and reliability; and self-correction
and recovery --- each restricted to
\texttt{cs.AI,\ cs.CL,\ cs.LG,\ cs.SE,\ cs.MA,\ cs.HC,\ stat.ML} and a
submission window of 2024-01 through 2026-07 (2023 foundational work is
handled separately as hand-verified anchors, §2.3, §5). Each thread was
capped at 250 results (a disclosed depth limit) and sorted by relevance.
Across the eight threads this returned 2,000 raw hits, 1,939 unique
after cross-thread deduplication by arXiv ID --- the low duplicate rate
(61 hits) indicates the threads are reasonably distinct rather than
redundant. A harvest-time relevance filter then required title+abstract
to contain at least one topic-signal term (e.g.~\emph{long-horizon},
\emph{multi-step}, \emph{episod-}, \emph{trajector-}, \emph{planning},
\emph{memory}, a named benchmark, \emph{credit assignment},
\emph{reflection}, \emph{replan-}, \emph{orchestrat-}, \emph{scaffold-})
so that the single most overloaded word in this literature --- ``agent''
--- could not by itself admit a paper; this dropped 176 hits (9.1\% of
the unique pool), leaving 1,763 seed candidates.

\textbf{Query bleed and reclassification.} Because ``agent'' and several
of the thread queries (especially \emph{credit assignment} and
\emph{reinforcement learning for agents}) also match a large classical
literature --- multi-agent reinforcement learning theory, robotic
manipulation and motion planning, autoregressive time-series/PDE
forecasting --- that this survey explicitly scopes out (classical HRL,
options, and robotic task-and-motion planning are treated as historical
anchors only, not as corpus members, following the boundary set in §1),
we ran an exploratory TF-IDF + truncated-SVD + \(k\)-means clustering
(\(k=18\)) over the seed pool purely as scaffolding for taxonomy design,
then reclassified every paper with cluster-based category defaults,
cross-cutting keyword rules, and an explicit LLM/agent-signal gate that
catches classical-literature bleed that the harvest-time filter's
broader terms let through (clusters that were agent-systems-specific by
construction --- e.g.~a GUI-agent cluster, an agent-memory-systems
cluster --- were exempted from this gate after we verified by inspection
that gating them produced false exclusions). This second-stage filter
excluded 344 further papers (19.5\% of the 1,763 seed candidates);
combined with the harvest-time filter, 520 of the 1,939 raw unique hits
(26.8\%) were judged off-topic and dropped over the two stages --- we
report this as an absolute number, following the recommendation that
corpora built around a term as overloaded as ``agent'' disclose bleed
quantitatively rather than assert its absence. A random sample of 30
kept papers and 55 excluded papers was manually re-checked against this
criterion; the estimated residual misclassification rate after both
filters is on the order of one in twenty, concentrated at the boundary
between LLM-driven and classical robotic/multi-agent-RL systems, which
is inherently fuzzy for a handful of transitional papers. This left
\textbf{1,419 seed papers}.

\textbf{Targeted supplement.} Reclassification exposed one severe gap
--- \texttt{foundations} (failure-mode theory, reliability/scaling,
oversight of long-running agents) held only 10 papers, well under the
threshold at which we would have folded it back into prose-only
treatment inside another section --- and two thinner spots
(cross-session memory persistence; credit-assignment theory). We ran 19
targeted arXiv queries against these directions (2024+,
\texttt{cs.*}/\texttt{stat.ML}), yielding 321 raw candidates; a stricter
secondary filter requiring both an explicit agent/agents token and a
foundations-or-long-horizon signal term dropped 193 (60.1\% --- a
markedly higher bleed rate than the seed harvest, consistent with these
being broader, less-targeted query terms such as ``reward hacking'' or
``scaling law'' that catch unrelated work on quantization, pretraining,
or physical-system forecasting), keeping \textbf{128 supplement papers},
tagged \texttt{source=supplement}. The supplement is recency-biased by
construction and is therefore excluded from the growth-timeline figure
(Figure 6, §9), which uses seed-only counts.

\textbf{Final corpus and enrichment.} The combined corpus is
\textbf{1,547 papers} (1,419 seed + 128 supplement) --- larger than our
initial \textasciitilde1,100--1,400 planning estimate; rather than force
the corpus down to a pre-registered size, we kept the full
bleed-filtered yield. Author and venue metadata were resolved via
batched OpenAlex lookups (by the arXiv DOI form), with a mandatory
arXiv-API fallback for the 180 rows (11.6\%) OpenAlex had not yet
indexed or indexed without authors; this achieved 100\% author coverage.
Because the corpus is overwhelmingly recent (58--77\% of most categories
was posted in 2026 alone, Table 1), citation counts would be near zero
for the majority of entries and are not a meaningful signal here; we do
not report or use them, and caution against reading corpus composition
as a proxy for impact anywhere in this paper.

\textbf{Single-label assignment is a simplification.} Each paper carries
exactly one \texttt{category} in the released corpus, but a large
fraction of this literature is genuinely multi-topic: a memory-augmented
agent evaluated on a new benchmark and trained with process rewards has
a legitimate claim to three categories. Our labels record the
\emph{primary} contribution --- the question the paper's own abstract
foregrounds --- not an exclusive membership claim. As a lower-bound
indicator of how often that primary-contribution call is a real judgment
rather than an obvious single fit, we counted how many of the six
categories' topic-keyword signals each paper's \emph{title alone}
triggers: \textbf{13.8\% of the corpus (213 papers) trips two or more}.
We report this as a keyword co-occurrence statistic, not a semantic
scope estimate --- it is a lexical measurement, not a claim about what a
domain expert would conclude, and it is itself an undercount in one
direction (a title like ``X: A Benchmark for Y'' trips both
\texttt{evaluation} and Y's signal even when the paper's unambiguous
primary contribution is the benchmark) and an overcount in another
(co-occurring keywords do not always mean co-equal contributions). We do
not extend this count to title-plus-abstract, where generic phrases like
``we evaluate'' fire the \texttt{evaluation} signal in nearly every
machine-learning abstract regardless of topic, making the resulting
number uninformative rather than merely noisy. The most common
title-level overlaps are memory+evaluation (29 papers, chiefly memory
benchmarks), memory+execution (23), and training+execution (22), which
is why §§4-7 cross-reference each other as often as they do. Readers
should treat per-category counts as a map of where each paper's
\emph{center of mass} sits, not as disjoint bins, and the release
includes full abstracts so that any alternative labeling can be
recomputed from the same corpus.

\textbf{Limitations.} This is a single-annotator, largely rule-based
corpus: classification followed cluster defaults, keyword rules, and an
explicit per-paper override list (never hand-edits to the underlying
data), all released as scripts. It under-represents unpublished
industrial agent engineering that never reaches arXiv,
non-English-language work, and any system described only in a blog post
or technical report rather than a paper. Benchmark leaderboard positions
move on a timescale of months; every specific benchmark number we cite
is dated at first mention rather than presented as a current ranking.

\textbf{Language calibration.} Throughout this paper we avoid unmeasured
strong language --- ``proves,'' ``monotone,'' ``law,'' ``first'' --- in
favor of ``observed,'' ``measured and reported,'' or ``tracks,''
reserving stronger language only for claims a cited paper's own
experiments establish directly. This applies with particular force to
the METR time-horizon trend (§7, §9): we treat its reported growth rate
as \emph{measured and reported}, not as a settled law, because the
metric's cross-model-era comparability and its generalization beyond
benchmark tasks are both contested rather than resolved.

\subsection{2.3 Taxonomy}\label{taxonomy}

We organize the corpus along one primary axis realized as six
categories, plus a second, prose-level axis that cuts across all six and
is this paper's main organizing device.

\textbf{Axis 1 --- what the paper is about}, the \texttt{category}
column reported in Table 1: \emph{planning \& decomposition} (§3: how
agents turn a task into a sequence of committable steps ---
\texttt{decompose}/\texttt{search}/\texttt{worldmodel} subcategories),
\emph{memory \& context management} (§4: how information persists within
and across a task --- \texttt{context}/\texttt{external} subcategories,
plus a weights tier discussed qualitatively, not separately tagged),
\emph{execution control \& recovery} (§5: the runtime loop that turns
plans into actions and recovers from failures ---
\texttt{loop}/\texttt{orchestration}/\texttt{recovery} subcategories),
\emph{training for long horizons} (§6: how models are trained to act
well over many steps --- \texttt{rl}/\texttt{supervision}
subcategories), \emph{evaluation \& measurement} (§7: how we know
whether any of the above works --- currently a single \texttt{benchmark}
subcategory, since every classified paper in it is benchmark work, plus
the critical literature that interrogates benchmarks throughout), and
\emph{foundations, limits \& safety} (§8: theory of degradation and
compounding error, and the oversight problem for agents that run for a
long time unsupervised --- not subdivided into subcategories). Table 1
reports per-category counts, the subcategory tags each category actually
carries, and the fraction of each category posted in 2026 alone.

\textbf{Table 1.} Corpus composition by category (1,547 papers total).

{\def\LTcaptype{none} 
\begin{longtable}[]{@{}
  >{\raggedright\arraybackslash}p{(\linewidth - 8\tabcolsep) * \real{0.1667}}
  >{\raggedright\arraybackslash}p{(\linewidth - 8\tabcolsep) * \real{0.1667}}
  >{\raggedleft\arraybackslash}p{(\linewidth - 8\tabcolsep) * \real{0.2222}}
  >{\raggedleft\arraybackslash}p{(\linewidth - 8\tabcolsep) * \real{0.2222}}
  >{\raggedleft\arraybackslash}p{(\linewidth - 8\tabcolsep) * \real{0.2222}}@{}}
\toprule\noalign{}
\begin{minipage}[b]{\linewidth}\raggedright
Category
\end{minipage} & \begin{minipage}[b]{\linewidth}\raggedright
Subcategory tags (n)
\end{minipage} & \begin{minipage}[b]{\linewidth}\raggedleft
Papers
\end{minipage} & \begin{minipage}[b]{\linewidth}\raggedleft
Of which supplement
\end{minipage} & \begin{minipage}[b]{\linewidth}\raggedleft
\% posted 2026
\end{minipage} \\
\midrule\noalign{}
\endhead
\bottomrule\noalign{}
\endlastfoot
Planning \& decomposition & decompose (162) · worldmodel (11) · search
(9) & 182 & 0 & 27\% \\
Memory \& context management & external (294) · context (103) & 397 & 21
& 72\% \\
Execution control \& recovery & orchestration (338) · recovery (245) ·
loop (1) & 584 & 0 & 45\% \\
Training for long horizons & rl (130) · supervision (37) & 167 & 14 &
62\% \\
Evaluation \& measurement & benchmark (114) & 114 & 0 & 58\% \\
Foundations, limits \& safety & not subdivided & 103 & 93 & 77\% \\
\end{longtable}
}

Subcategory tags are a corpus column for five of six categories;
\texttt{evaluation} currently resolves to a single tag
(\texttt{benchmark}) because every classified paper in it is benchmark
work, and \texttt{foundations} was not given a subcategory split (§8's
internal organization --- compounding error, goal drift/misalignment,
oversight --- is a prose-level grouping, not a tagged column, matching
how Axis 2 below is handled).

\begin{figure}
\centering
\includegraphics[width=0.95\linewidth,height=\textheight,keepaspectratio,alt={Figure 1: Taxonomy grid -- six categories (rows) crossed with the horizon-bearing-locus axis (columns). Each cell names representative systems drawn from the corpus and discussed in the corresponding section. The foundations/within-context cell is empty by construction: degradation and oversight phenomena presuppose a task that already exceeds one forward pass.}]{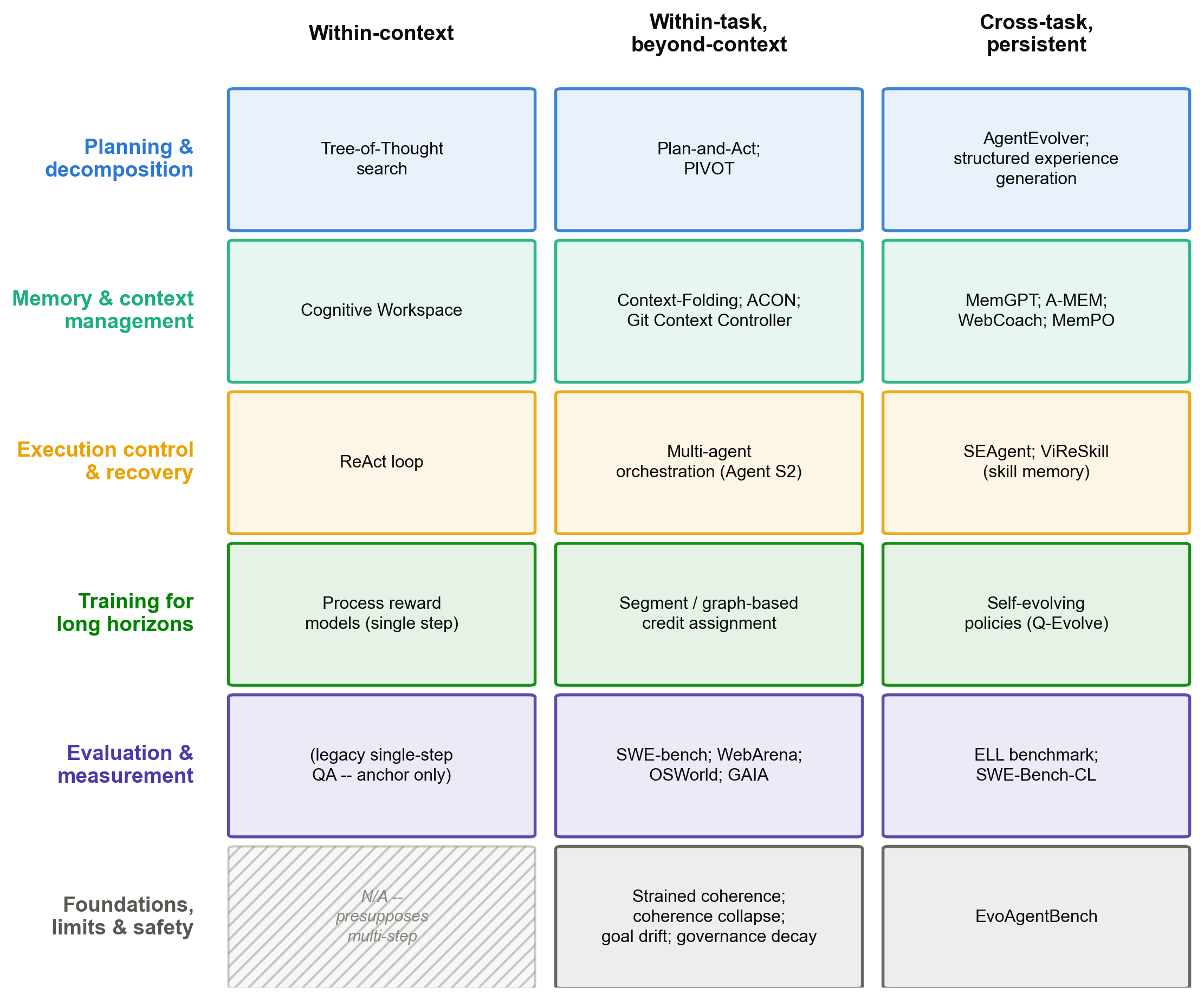}
\caption*{Figure 1: Taxonomy grid -- six categories (rows) crossed with
the horizon-bearing-locus axis (columns). Each cell names representative
systems drawn from the corpus and discussed in the corresponding
section. The foundations/within-context cell is empty by construction:
degradation and oversight phenomena presuppose a task that already
exceeds one forward pass.}
\end{figure}

Figure 1 renders Axis 1 crossed with Axis 2 as a grid, with
representative systems per cell drawn from the corpus and each discussed
in the corresponding section below. Seventeen of the eighteen cells are
populated: only the \texttt{foundations}/within-context cell is empty,
because degradation and oversight problems are, definitionally,
phenomena of tasks that already exceed a single forward pass. The
cross-task-persistent column is the sparsest in the literature but is
populated in every other row, which is worth noting explicitly --- the
impulse to let capability accumulate across episodes is not confined to
the memory literature where it is most visible, but appears as a
distinct minority thread inside planning (§3), execution (§5), training
(§6), evaluation (§7), and the self-evolution work discussed in §8.
Figure 2 shows the same six categories as a semantic map (TF-IDF
projection with per-category density contours), which makes visible that
\texttt{execution} and \texttt{planning} interpenetrate heavily ---
consistent with the two sharing much of their harness-engineering
literature --- while \texttt{training} forms the most topically
separated cluster.

\begin{figure}
\centering
\includegraphics[width=0.95\linewidth,height=\textheight,keepaspectratio,alt={Figure 2: Semantic map of the corpus (TF-IDF + SVD + t-SNE projection, axes are arbitrary embedding dimensions) with per-category density contours and direct labels. n=1,547.}]{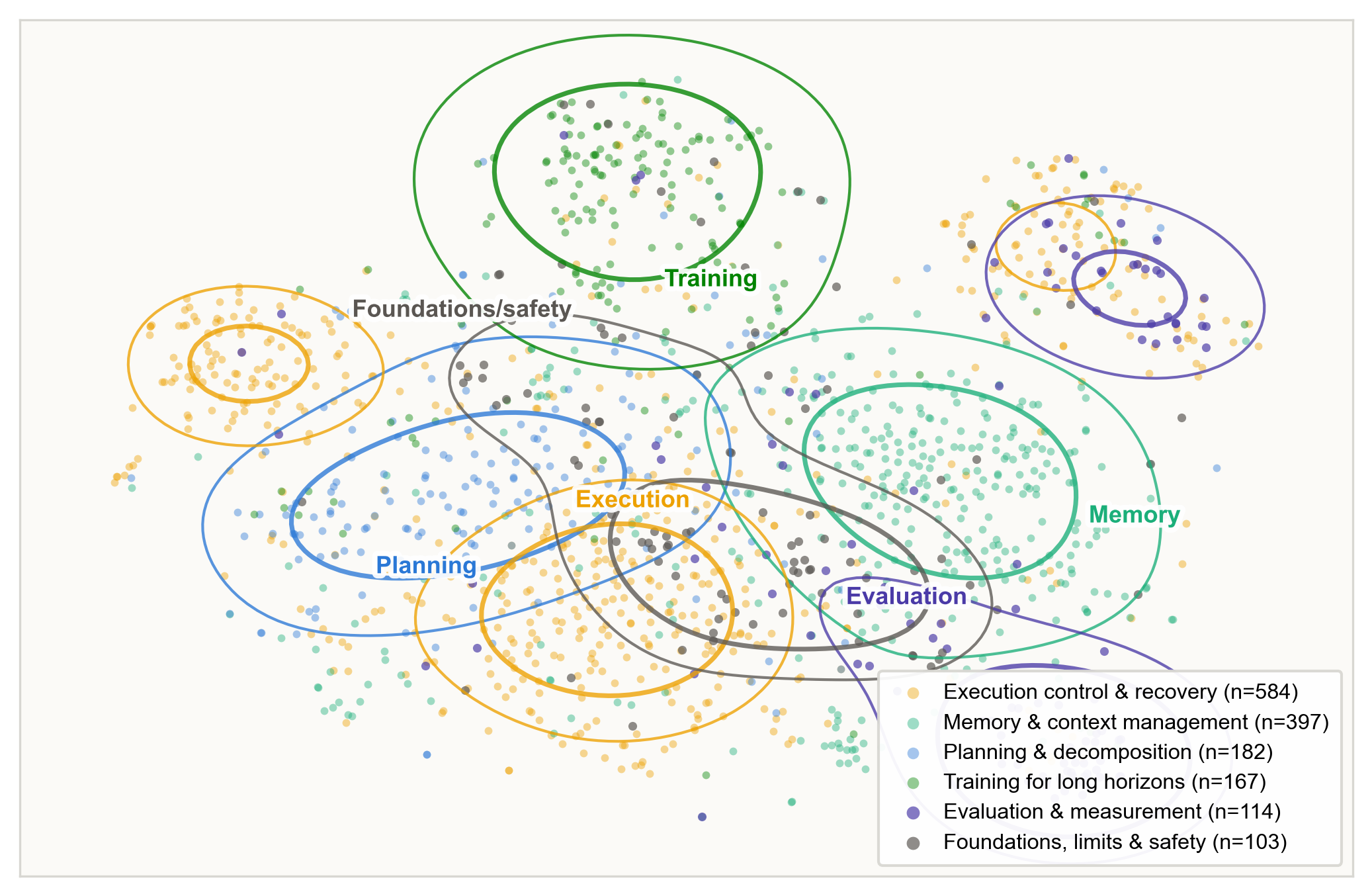}
\caption*{Figure 2: Semantic map of the corpus (TF-IDF + SVD + t-SNE
projection, axes are arbitrary embedding dimensions) with per-category
density contours and direct labels. n=1,547.}
\end{figure}

\textbf{Axis 2 --- where the horizon is carried}, a prose-level
distinction we apply within every technical section rather than as a
corpus column: \textbf{within-context} (the task's full history fits in
one forward pass --- ``long-horizon'' here means many steps, not
literally beyond-context), \textbf{within-task-beyond-context} (the step
count exceeds one context window, but a harness --- external memory,
summarization, sub-agent handoff --- keeps the attempt inside one
continuous episode), and \textbf{cross-task-persistent} (information or
skill survives across distinct episodes or sessions --- memory banks,
fine-tuned weights, accumulated skill libraries --- so horizon is
extended by accumulation over time rather than within a single attempt).
Every technical section below is, in part, organized by where along this
axis its methods sit, because the engineering problem --- and the
failure modes --- differ sharply across the three: within-context
methods fail by attention/utilization degradation (§4, §8);
within-task-beyond-context methods fail by information loss at the
harness boundary; and cross-task-persistent methods fail by interference
or staleness in what was retained.

\subsection{2.4 Positioning vs.~existing
surveys}\label{positioning-vs.-existing-surveys}

Several recent surveys cover pieces of this territory. Planning surveys
such as \citep{huang2024understanding} and \citep{tantakoun2025llms} map
the space of LLM planning strategies; memory surveys such as
\citep{zhang2024survey} and its 2026 update \citep{huang2026rethinking}
chart agent memory mechanisms in detail; GUI/web-agent surveys such as
\citep{zhang2024large}, \citep{nguyen2024gui}, and
\citep{tang2025survey} cover a single embodiment of the execution
problem; and an agentic-reinforcement-learning survey
\citep{zhang2025landscape} covers one training paradigm in §6's
territory. Each is organized around a single component of the pipeline.
This survey's delta is to organize instead around \textbf{horizon} as
the cross-cutting axis: rather than asking ``what memory mechanisms
exist'' or ``what planning strategies exist'' in isolation, we ask, for
every stage of the pipeline, \emph{what breaks first as the required
horizon grows, and what compensates} --- which is why planning, memory,
execution, training, evaluation, and foundations are treated as one
continuous argument (§§3--8) rather than as independent surveys stitched
together, and why the critical/diagnostic literature (benchmark
critiques, failure-mode analyses, self-correction skepticism such as
\citep{kamoi2024when}) is carried as a first-class thread through every
section rather than segregated into its own critique chapter.

This survey is also a deliberate companion to a prior survey of
recursive self-improvement in AI \citep{chen2026rsi}, with which it
shares harvest infrastructure and taxonomy methodology but not scope:
that survey covers loops in which a system improves itself --- its
outputs, its training data, its evaluator, or the research process ---
as an end in itself; this survey covers training and engineering aimed
specifically at extending how long a task a system can carry out, and
treats self-improvement loops only where they are instrumental to that
end (§6). Readers interested in \emph{why} models are trained on their
own generated data, rather than \emph{how far a single trained policy
can be pushed on long tasks}, should consult the companion survey; we
flag the boundary again where it becomes concrete, in §6.

\section{3. Planning \& Decomposition}\label{planning-decomposition}

Planning is the first place a long-horizon task must be made tractable:
before any action is taken, something has to turn ``resolve this issue''
or ``book this itinerary'' into a sequence of committable steps. The
literature spans a spectrum from committing to a full plan before any
execution, through interleaving one planning step with one execution
step, to explicit search over multiple candidate continuations --- and,
as the corpus grew, it became clear that where a system sits on this
spectrum tracks how uncertain its environment is, not how sophisticated
its model is. Plan-first decomposition is data-efficient and
interpretable precisely when the environment is predictable enough that
a plan made at \(t=0\) still holds at \(t=k\); as uncertainty grows,
that assumption breaks, and the field's center of mass has moved toward
interleaved and search-based alternatives (this section,
\texttt{decompose} \(n=162\), \texttt{search} \(n=9\),
\texttt{worldmodel} \(n=11\) of the taxonomy's \texttt{planning}
category).

\begin{figure}
\centering
\includegraphics[width=0.9\linewidth,height=\textheight,keepaspectratio,alt={Figure 3: The planning strategy spectrum -- plan-then-execute, interleaved, search-based, and world-model-based lookahead approaches, positioned against increasing environment uncertainty and action-space size.}]{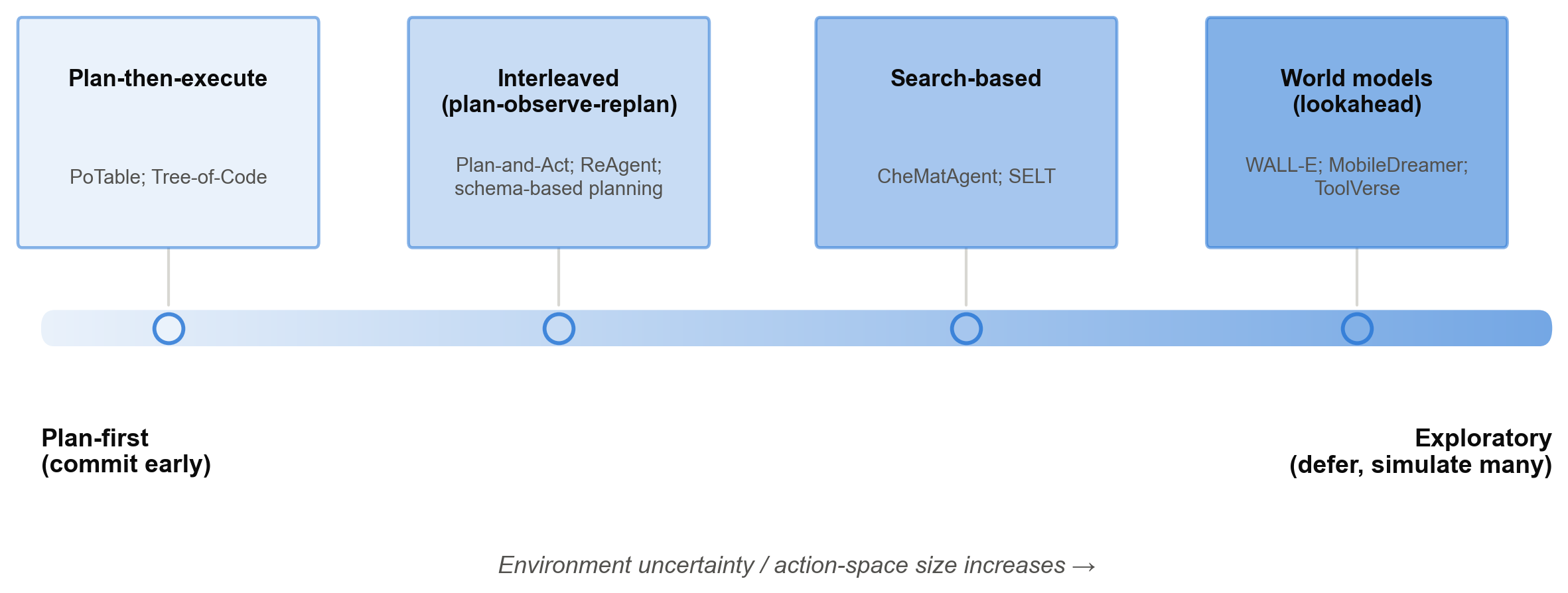}
\caption*{Figure 3: The planning strategy spectrum -- plan-then-execute,
interleaved, search-based, and world-model-based lookahead approaches,
positioned against increasing environment uncertainty and action-space
size.}
\end{figure}

\textbf{Plan-then-execute and its failure mode.} The idea that a long
task should be attacked by decomposing it into temporally extended
sub-behaviors long predates language models: the options framework
formalized temporal abstraction over sub-policies in reinforcement
learning \citep{sutton1999between}, and feudal reinforcement learning
organized control into a manager hierarchy assigning sub-goals to
workers \citep{dayan1992feudal} --- the structural ancestor of the
planner/executor splits that recur throughout this section. Two
2022-2023 systems carried the pattern into the LLM setting from opposite
directions: LLM+P routes planning through a classical symbolic planner,
using the language model to translate the problem into a formal
specification rather than to plan directly \citep{liu2023llmp}, while
SayCan grounds an LLM's proposed steps in what a robot can actually
execute by weighting them against learned affordances
\citep{ahn2022saycan} --- an early statement of this section's recurring
lesson that a plan is only as good as its feasibility check. The
classical recipe --- decompose the task into sub-goals, then execute
each --- remains the default in tool-use and code-generation settings
where the action space is well-specified: PoTable applies staged
plan-then-execute reasoning to table QA \citep{mao2024potable}, and
Tree-of-Code consolidates an LLM agent's actions into a unified
code-based action space precisely to make each planned step verifiable
before it executes \citep{ni2024tree}. But the recipe's central risk is
committing early to a plan the agent cannot actually carry out:
Plan-and-Act starts from the observation that LLMs are not inherently
trained to produce accurate plans, and trains a Planner separately from
an Executor using synthetic data that annotates ground-truth
trajectories with the feasible plans that would have produced them ---
teaching plan generation directly, rather than hoping a single model's
general capability transfers to it --- which recovers much of the gap on
long-horizon web-navigation tasks \citep{erdogan2025plan}. PIVOT goes
further and treats the plan itself as an object to be refined against
execution feedback rather than committed to once --- plan, inspect, and
evolve trajectories through repeated environment interaction ---
precisely because plans generated before any grounding routinely violate
constraints or omit infeasible-action checks that only surface once
execution starts \citep{zhang2026pivot}. The same lesson appears from
the harness-design side: harnesses that decompose more aggressively are
not uniformly better, because more elaborate decomposition can itself
introduce misalignment between the guidance a step receives and what
execution actually needs, so the right amount of decomposition is a
property of the task's uncertainty, not a free efficiency gain
\citep{wang2026harnesses}.

\textbf{Interleaving planning with execution.} The alternative --- plan
one step, observe, replan --- trades a priori coherence for grounding.
PROMST shows that multi-step tasks need feedback- and heuristic-informed
prompt optimization that single-step prompt tuning does not require,
precisely because the impact of an individual step is hard to evaluate
in isolation from what happens next \citep{chen2024prompt}; ReAgent
makes the interleaving explicit and reversible, adding backtracking to
multi-hop QA specifically because irreversible chain-of-thought
accumulates errors across hops that only a revisable trajectory can undo
\citep{zhao2025reagent}. A sharp empirical question is exactly how much
interleaving is warranted: directly testing full-horizon planning
(commit to a complete plan, then execute) against single-step-horizon
planning (interleave every action with fresh reasoning) on data-centric
tool-calling tasks, one study finds full-horizon planning with on-demand
replanning matches step-by-step accuracy across the depths, breadths,
and robustness levels tested --- at 2-3x fewer tokens --- indicating the
interleaved default's assumption that eager step-wise monitoring is
necessary for adaptability does not hold for well-defined data-centric
tasks, even though it is treated as the safe default
\citep{otani2026agents}. This is a challenge to committing to
interleaving by default, not a demonstration of the
predictability-conditioned split this section's opening claims. A
cleaner test of that specific split is still open. At the far end of
this axis, when the action space itself becomes too large to plan over
explicitly (open-ended real-world environments), one line of work argues
the representation planning happens over must change --- from planning
with concrete actions to planning with higher-level schemas that
compress the combinatorially exploding action space back down to
something a plan can range over \citep{xu2025cognitive}.

\textbf{Search-based planning.} A third family keeps multiple candidate
continuations alive and searches among them rather than committing to
one. The anchor for this family is Tree of Thoughts, which generalizes
chain-of-thought from a single linear trace into a tree of partial
solutions the model can evaluate and backtrack over \citep{yao2023tree};
the within-window variant remains the reference point against which
later, tool- and environment-grounded search methods are described. In
our 2024-2026 corpus this is the smallest slice (\texttt{search},
\(n=9\)) but a conceptually distinct one: CheMatAgent integrates
tree-search over 137 external chemistry tools specifically because a
single greedy tool-call sequence is unreliable when the right tool
depends on intermediate results not known in advance
\citep{wu2025chematagent}, and SELT combines self-evaluation with tree
search over decomposed sub-tasks so that the search process itself, not
just the final answer, is guided by a learned value signal
\citep{wu2025selt}. The progression is worth noting: Tree of Thoughts
searches over \emph{reasoning states} inside one context, while its
successors search over \emph{action sequences} whose outcomes only an
environment can reveal --- which moves the bottleneck from search
strategy to the cost and fidelity of evaluating each branch.
Search-based planning is the most compute-hungry point on the spectrum
and the literature increasingly asks whether the compute is well spent
--- a question §7 returns to when discussing benchmark protocol
validity.

\textbf{World models for lookahead.} A fourth thread --- closely related
to search, since search needs something to search \emph{over} --- asks
agents to maintain an explicit predictive model of the environment and
plan by rolling it forward. WALL-E aligns a world model to real
environment rules via rule learning specifically because an ungrounded
world model's predictions drift from what the environment actually does
over a long rollout, and shows this alignment step recovers much of the
value of model-based planning that an ungrounded model loses
\citep{zhou2024wall}. This drift is a recurring finding: a comparison of
memory mechanisms in world models finds that the effective planning
horizon of transformer-based world models is bounded by their effective
memory span, and that perceptual drift compounding across a long
imagined rollout breaks the loop-closure a plan depends on ---
connecting this thread directly to the memory literature of §4 and the
compounding-error theory of §8 \citep{j2025memory}. Applied work shows
both the promise and the constraint: MobileDreamer builds a generative
sketch world model specifically to let a mobile GUI agent forecast
action outcomes before committing to them, addressing agents that are
otherwise purely reactive to the current screen and therefore blind on
long-horizon tasks \citep{cao2026mobiledreamer}, while ToolVerse scales
agentic-RL environments precisely because agents that reason well in
compact, well-defined scenarios do not automatically transfer that
lookahead ability to large, diverse, dynamic ones
\citep{zhou2026toolverse}.

\textbf{Planning that persists across tasks.} All four families above
operate within a single attempt. A smaller thread asks whether the
planning capability itself can accumulate across episodes --- the
cross-task-persistent point on Axis 2 (§2.3) as it applies to planning
rather than to memory. AgentEvolver treats the expensive parts of
building an agent --- constructing task datasets and exploring for
useful trajectories --- as things the system should bootstrap for itself
rather than have hand-supplied per task, on the argument that manually
constructed datasets and undirected RL exploration are what make agent
development costly and sample-inefficient in the first place
\citep{zhai2025agentevolver}. SEAM takes a narrower and more
architecturally explicit route: rather than retrieving past experience
by similarity at inference time --- which it argues introduces noise and
latency --- it trains a small executor-specific adapter that
\emph{generates} a structured, instance-tailored experience entry in one
forward pass for a frozen executor, storing the accumulated experience
in the adapter's parameters and improving it after deployment from
logged successful trajectories \citep{li2026beyonda}. The two mark the
ends of a small design space --- bootstrapping the whole training
pipeline versus compiling reusable experience into a lightweight side
module --- and both make explicit an assumption the rest of this section
leaves implicit: that a plan's value can outlive the episode that
produced it.

\textbf{Assessment.} Across the four within-episode families, the same
pattern recurs: methods that commit to a representation of the future
early (a full plan, an unrolled world-model trajectory) buy coherence at
the price of fragility to whatever the plan did not anticipate, while
methods that defer commitment (interleaving, search) buy robustness at
the price of compute and sometimes at the price of the very global
coherence long-horizon tasks need. No single point on the spectrum
dominates; which point is appropriate is closest to being a measurable
property of the environment's uncertainty and the action space's size,
not a fixed methodological preference --- which is why the field's
center of mass has moved from plan-then-execute defaults toward
interleaved and search-augmented methods as it has moved from short,
well-specified benchmarks toward open-ended, long-horizon ones (§7). The
cross-task thread is the least developed of the five and the one where
the boundary with §4 is thinnest: once planning experience is stored and
reused, the engineering problem starts to look like a memory problem,
which is one reason the two categories interpenetrate as heavily as
Figure 2 shows. The plans this section describes are only as good as the
information available when they are made and revised, which is the
memory and context-management problem taken up next.

\section{4. Memory \& Context
Management}\label{memory-context-management}

If planning decides what to do next, memory decides what information
that decision is made with --- and this is where the
long-horizon/long-context/long-term-memory distinction of §2.1 does the
most work. We organize this section by where information is carried:
\textbf{in-context} (part of the live prompt), an \textbf{external
store} (retrieved on demand), or \textbf{weights} (compiled into the
policy via training) --- three points on a persistence-fidelity
trade-off rather than three unrelated mechanisms (\texttt{context}
\(n=103\), \texttt{external} \(n=294\) of the \texttt{memory} category;
the size gap itself is a finding, discussed in the Assessment below ---
the weights tier is not separately tagged in the corpus, discussed
qualitatively via the \texttt{training} category's overlap with this
section).

\begin{figure}
\centering
\includegraphics[width=0.9\linewidth,height=\textheight,keepaspectratio,alt={Figure 4: The memory persistence-fidelity trade-off across three tiers -- context, external store, and weights -- with representative systems and relative fidelity/persistence for each.}]{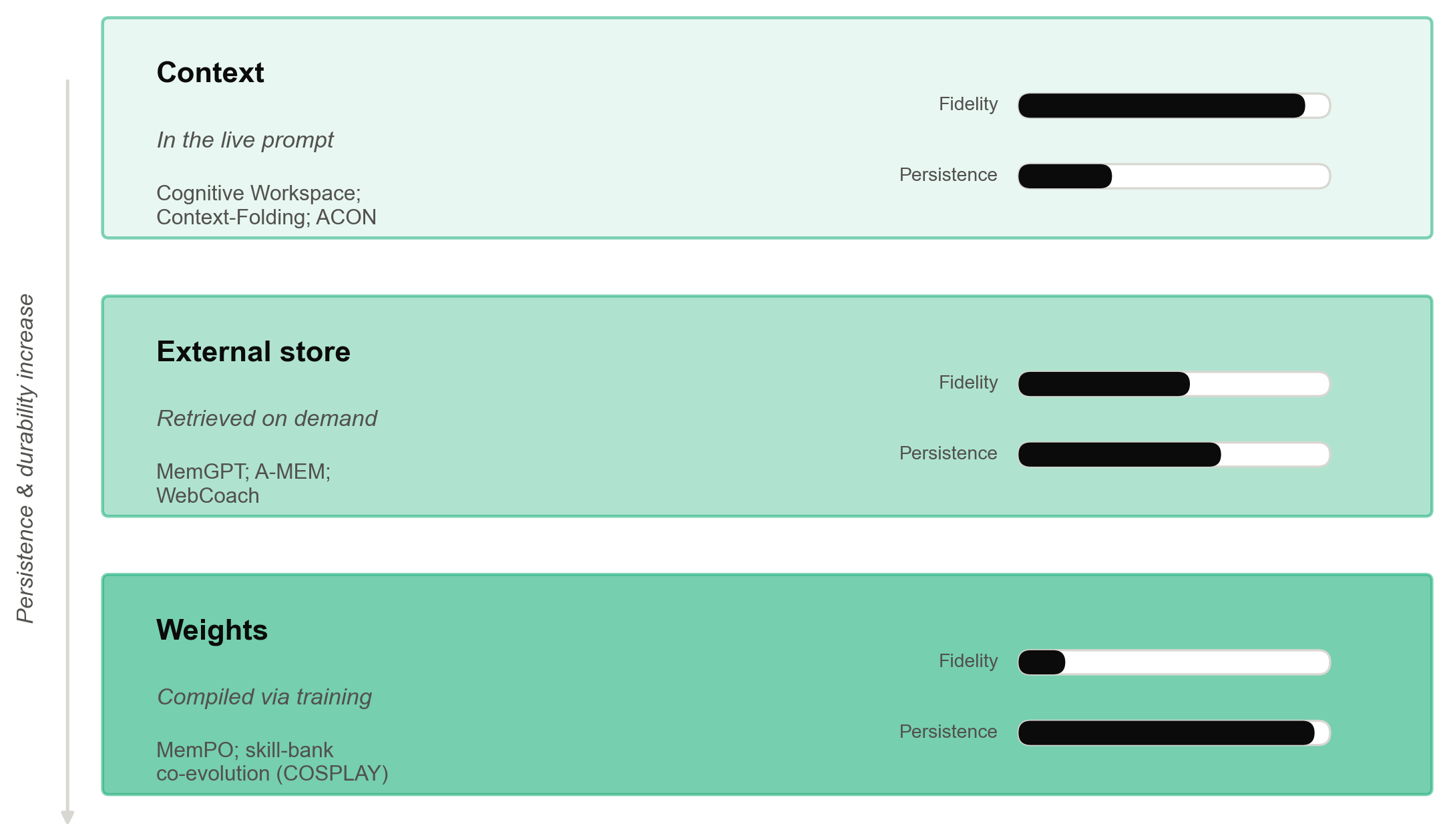}
\caption*{Figure 4: The memory persistence-fidelity trade-off across
three tiers -- context, external store, and weights -- with
representative systems and relative fidelity/persistence for each.}
\end{figure}

\textbf{In-context management.} The simplest way to preserve information
is to keep it in the prompt, but a long-horizon task eventually produces
a trajectory longer than any context window, and even within budget,
models do not use long contexts uniformly --- information placed away
from the ends of the context is used less reliably than information at
the edges \citep{liu2024lost}. Agent-specific work treats this as an
active management problem rather than a passive capacity limit:
Cognitive Workspace frames context management as active memory
management with a finite functional budget, directly analogous to human
working memory constraints, rather than as a fixed-size buffer to be
filled and discarded \citep{an2025cognitive}; Git Context Controller
manages an agent's context the way a version control system manages a
codebase --- commits, branches, and merges of context state --- because
ad hoc truncation loses exactly the information a later step turns out
to need \citep{wu2025git}. Several 2026 systems attack the
trajectory-length problem directly: Context-Folding scales a
long-horizon agent by folding completed sub-trajectories out of the live
context once they are resolved \citep{sun2025scaling}, and ACON
optimizes context compression specifically for the long-horizon setting,
showing that generic summarization compresses away details a later step
needs, so the compression policy itself has to be aware of what
downstream steps will require \citep{kang2025acon}. The failure mode
this thread converges on has recently been given a name --- ``context
rot,'' diagnosed and partially mitigated in long-horizon search settings
\citep{xia2026diagnosing} --- and a blunt summary: context management is
load-bearing for whether an LLM agent's plan survives contact with a
long trajectory, not an implementation detail behind it
\citep{mehta2026plans}.

\textbf{External memory.} The alternative to carrying everything
in-context is to write information to an external store and retrieve it
when relevant. This is by far the largest subcategory in the corpus
(\texttt{external}, \(n=294\) vs.~\texttt{context}'s 103), and the
operating system metaphor that popularized the idea --- MemGPT, treating
an LLM's limited context as working memory backed by a paged external
store, the way an operating system backs limited RAM with disk
\citep{packer2023memgpt} --- still organizes how most of this literature
frames the problem. A second anchor supplies the other half of the
design vocabulary: Generative Agents paired a timestamped memory stream
with a retrieval function scoring entries on recency, importance, and
relevance, plus a reflection step that periodically synthesized
higher-level inferences from raw observations \citep{park2023generative}
--- establishing that an external store needs a \emph{write} and
\emph{consolidation} policy, not only a read path, a point the
interference findings below return to. Recent systems differ mainly in
what the external store looks like and how retrieval is scoped: A-MEM
structures agentic memory as a network of interlinked notes rather than
a flat log, so that retrieval can follow associative links instead of
similarity search alone \citep{xu2025mem}; graph-based memory
architectures generalize this further, arguing a single flat vector
store under-represents the relational structure a long-running agent
accumulates \citep{yang2026graph}. Whether memory survives across
sessions, not just within one long episode, is this subcategory's
clearest engineering marker of the cross-task-persistent point on Axis 2
(§2.3): WebCoach gives web agents self-evolving guidance carried across
sessions specifically because within-session learning is invisible to
the next session's fresh start \citep{liu2025webcoach}, and a
scale-conditioned evaluation protocol finds that fixed-snapshot accuracy
hides a real failure mode: reliability degrades as merely
\emph{irrelevant} sessions accumulate around the actually-relevant
evidence, well before the store approaches any hard capacity limit ---
the protocol reports the specific accumulation scale at which
reliability crosses below target as a distinct diagnostic in its own
right \citep{shao2026when}. A closely related but distinct problem is
interference from evidence that is not irrelevant but \emph{outdated}: a
benchmark evaluating memory under multi-target interference in
long-horizon agent systems finds consistently low accuracy across seven
representative systems (vanilla long-context LLMs, RAG, and
memory-augmented frameworks; average 27.9\%), with questions that
require aggregating multiple relevant facts singled out as especially
hard even against that already-low baseline, and degradation compounding
further as facts are revised or updated by later context ---
interference from \emph{change over time}, not merely from how much is
stored, is the bottleneck \citep{lee2026minteval}, and forgetting is
correspondingly being explored as a first-class design target rather
than a failure to prevent --- biologically-inspired forgetting
mechanisms compress or discard low-value memories specifically to keep a
growing store's retrieval useful \citep{wei2026fademem}.

\textbf{Compression and distillation into weights.} The third tier folds
information into the model itself rather than an external artifact ---
the most persistent but least flexible point on the trade-off, since
updating it requires training rather than a write to a store. MemPO
frames this as self-memory policy optimization for long-horizon agents,
training the policy itself via reinforcement learning to autonomously
decide what to retain or summarize based on downstream usefulness ---
the skill of curation becomes a trained capability baked into the
weights, rather than an external module the policy merely calls
\citep{li2026mempo}; co-evolving decision and skill-bank agents blend
the tiers explicitly, pairing a fast-changing external skill bank with a
more slowly updated decision policy so that frequently reused skills
eventually migrate from retrieval into the policy itself
\citep{wu2026co}. This tier is the most durable against context-window
limits by construction, and correspondingly the least auditable --- a
concern §8 returns to.

\textbf{A cross-cutting risk: memory as an attack surface.} Because
external and weight-level memory both persist information an attacker
does not control at write time, several 2026 papers treat agent memory
itself as a security boundary rather than only a capability:
conversational interaction can plant a stealthy trojan directly into an
agent's stored memory \citep{wang2026hijacking}, and a forensic
trajectory signature for detecting memory-poisoning after the fact
reaches high discrimination in a preregistered evaluation, but a
follow-up finds the same signature also fires on benign memory-grounded
behavior --- a reminder that a detector's headline accuracy and its
practical false-positive rate under realistic use are separate questions
this literature is only beginning to ask \citep{leong2026forensic}. This
connects the memory literature directly to the oversight problem of §8:
a memory store is exactly the kind of persistent, agent-controlled state
that is hard for a human overseer to audit continuously.

\textbf{Assessment.} The \texttt{external}-to-\texttt{context} size
ratio in the corpus (294 vs.~103) is itself informative: as of this
writing, the field has moved decisively toward treating memory as a
storage-and-retrieval engineering problem rather than a
context-budgeting one, likely because external stores scale
independently of a fixed context window while in-context management
schemes are ultimately still bounded by it. But the persistence-fidelity
trade-off runs through all three tiers regardless of implementation:
context is faithful but bounded, external stores are unbounded but
suffer retrieval and interference costs that grow with scale, and
weights are durable but require training to update and are the hardest
tier to audit or correct after the fact. None of the three eliminates
the trade-off; they only choose where on it to sit. Whichever tier a
system chooses, the resulting information still has to be acted on
inside a runtime loop that can fail and needs to recover --- the subject
of §5.

\section{5. Execution Control \&
Recovery}\label{execution-control-recovery}

Planning and memory supply what to do and what to do it with; execution
control is the runtime loop that actually does it, and --- the recurring
claim of this section --- how well a long task goes depends more on that
loop's ability to notice and recover from failure than on the per-step
quality of the underlying model. This is the corpus's largest category
(\texttt{execution}, \(n=584\)), organized into three subcategories that
sit at increasing scale: the single-agent act--observe \textbf{loop}
that turns a model into an agent at all, \textbf{orchestration} across
multiple agents or sub-agents (\(n=338\), the dominant mode by volume),
and \textbf{recovery} via self-correction and replanning when the loop's
output is wrong (\(n=245\)).

\textbf{The loop itself.} The interleaved reason-act-observe cycle ---
think, take an action, read back the environment's response, repeat ---
is now so assumed a substrate for LLM agents that the corpus contains
almost no papers that treat the loop itself as a subject rather than
infrastructure (\texttt{loop}, \(n=1\)); the pattern's near-total
absorption into ``how agents just work'' is itself informative about how
settled this layer has become since ReAct first proposed it as an
explicit alternative to acting without intermediate reasoning traces
\citep{yao2022react}. What the corpus does discuss is how to make the
loop \emph{safe} to run for a long time unsupervised: a
practitioner-facing guide to architecting resilient plan-then-execute
agents treats the loop's exception paths --- what happens when a tool
call fails, times out, or returns something the plan did not anticipate
--- as first-class design surface rather than an afterthought bolted
onto a working happy path \citep{f2025architecting}.

\textbf{Orchestration across agents.} The dominant way the field has
scaled beyond a single loop is horizontally: split a task across
multiple agents or sub-agents rather than deepen one agent's own loop.
Two anchors bracket what ``delegation'' came to mean: Toolformer had a
model learn, in a self-supervised way, when to call an external API
mid-generation --- delegation to a \emph{tool}
\citep{schick2023toolformer} --- while HuggingGPT used a language model
as a controller that plans a task and dispatches subtasks to other
specialist models, delegation to a \emph{model}
\citep{shen2023hugginggpt}. Present-day orchestration generalizes the
latter: the delegate is another agent with its own loop. This ranges
from domain-specialized multi-agent frameworks --- for requirements
engineering \citep{jin2025iredev}, sentiment analysis
\citep{xu2025sentimm}, penetration testing
\citep{luong2025xoffense, lin2025comparing}, optical-network operations
\citep{zhang2025generative}, and incident response with an explicit
deterministic-decision-support goal \citep{drammeh2025multi} --- to
general-purpose compositional frameworks for a single embodiment, such
as Agent S2's generalist-specialist split for computer-use agents
\citep{agashe2025agent} and OmegaUse's general-purpose GUI agent
\citep{zhang2026omegause}. A second thread treats orchestration itself,
not any one domain, as the object of study: the first large-scale
empirical study of testing practices across 39 open-source agent
frameworks and 439 agentic applications finds testing effort
concentrated almost entirely on the deterministic scaffolding around a
model --- tools and workflows absorb over 70\% of testing effort ---
while the model-driven planning component itself receives under 5\% and
prompts under 1\%, an inversion that leaves exactly the least
deterministic, most agentic part of the system least tested
\citep{hasan2025empirical}; a two-dimensional framework separates an
agent's cognitive function from its execution topology specifically
because neither axis alone disambiguates architecturally distinct
systems --- the same orchestration topology can implement patterns with
very different failure modes \citep{huang2026two}. The governance
literature has started treating agentic deployment as a policy object in
its own right, from two different angles: ``Governing AI Agents''
applies the economic theory of principal-agent problems to argue that
conventional governance tools --- monitoring, incentive design,
enforcement --- may not transfer to AI agents that act at a speed and
opacity ordinary agency relationships did not anticipate
\citep{kolt2025governing}, while the AI Agent Index, a public database
of deployed agentic systems, finds developers document capabilities and
applications far more thoroughly than they document safety and
risk-management practices \citep{casper2025ai} --- a disclosure gap this
section shares with the oversight discussion in §8.

\textbf{Recovery: self-correction and replanning.} The third subcategory
is the loop's error-handling layer --- noticing that an action or a
reasoning step was wrong and repairing it before it compounds. The
pattern's anchor is Reflexion, which converts a failed attempt into
natural-language self-feedback stored in an episodic buffer and
conditions the next attempt on it --- verbal reinforcement in place of a
gradient update \citep{shinn2023reflexion}. Early optimism that models
could reliably self-correct their own reasoning without external
feedback was tempered by a widely cited finding that intrinsic
self-correction --- correction using only the model's own signal, with
no external verifier or ground truth --- does not reliably improve
reasoning accuracy and can degrade it
\citep{huang2024cannotselfcorrect}; the corpus's own trajectory traces
this tension rather than resolving it. Several 2024-2025 papers argue
for a more qualified intrinsic capability under narrower conditions ---
self-correction of single-utterance perturbed reasoning
\citep{sam2025language}, correction with explicit key-condition
verification \citep{wu2024large} --- while a 2025 decomposition of
self-correction into detection, localization, and correction
sub-capabilities surfaces a genuinely counter-intuitive finding it calls
the ``accuracy-correction paradox'': the weaker model in a three-model
comparison (GPT-3.5, 66\% base accuracy) corrects its own errors
intrinsically at a \emph{higher} rate (26.8\%) than the strongest model
(94\% base accuracy, 16.7\%), and error-detection rate does not predict
correction success either. The proposed explanation --- an ``Error Depth
Hypothesis'' --- is that stronger models make fewer but structurally
deeper errors that intrinsic self-correction is specifically bad at
reaching, which would mean the anchor finding above is not a capability
gap current models simply haven't crossed yet, but a pattern that could
get \emph{worse}, not better, as models improve
\citep{li2025decomposing}. Where self-correction is externally grounded,
the picture is more positive: CSC-SQL uses corrective self-consistency,
checking multiple candidates against each other rather than relying on
one model's unaided judgment, to improve text-to-SQL reliability
\citep{sheng2025csc}, and SHIELDA gives agentic workflows structured
exception handling analogous to a programming language's exception
model, so that failures are caught and routed rather than silently
propagated \citep{jingwen2025shielda}. At the level of a full trajectory
rather than a single step, task-decoupled planning replaces one
monolithic reasoning history spanning every sub-task with a directed
acyclic graph of scoped sub-goals, each with its own confined context,
precisely because a shared history lets an error made on one sub-task
propagate into otherwise-unrelated decisions and makes recovery
correspondingly expensive to localize \citep{li2026beyond}.

\textbf{Recovery that accumulates.} A distinct move within the recovery
literature is to treat each failure-and-repair episode as a durable
asset rather than a one-off correction --- the cross-task-persistent
locus (§2.3) applied to execution. The anchor here is Voyager, which
accumulated a library of executable skills in an open-ended Minecraft
setting so that capabilities acquired solving one task became available
for the next \citep{wang2023voyager}. ViReSkill pairs vision-grounded
replanning with a skill memory: when execution fails, the replanner
generates a new action sequence conditioned on the current scene, and
when it succeeds, the resulting plan is retained for reuse --- so the
system's competence on a task family grows with the number of failures
it has worked through \citep{kagaya2025vireskill}. SEAgent makes the
same bet on a larger scale for computer-use agents, letting an agent
autonomously master unfamiliar software through iterative
trial-and-error on auto-generated tasks, precisely because human-labeled
demonstrations do not exist for novel or specialized applications
\citep{sun2025seagent}. Both depend on something §5's within-episode
recovery methods do not need: a reliable judgment about which past
attempts were actually successes, since a skill memory that accumulates
plausible-looking failures degrades rather than improves --- which
routes the problem back to the evaluation question of §7.

\textbf{Assessment.} The subcategory sizes are themselves a claim:
orchestration outnumbers recovery roughly 1.4 to 1, and both dwarf the
single-loop subcategory, suggesting the field has invested far more
engineering effort in \emph{scaling out} (more agents, more structure
around them) than in \emph{hardening} any one agent's own
error-correction --- even though the self-correction evidence above
suggests hardening is where the harder unsolved problem sits. This is
consistent with this section's opening claim: what separates a system
that completes a long task from one that does not is decreasingly a
property of the underlying model and increasingly a property of the
harness wrapped around it --- how failures are caught, how many agents
share the load, and how aggressively the system attempts to fix its own
mistakes before they compound. The cross-task-persistent accumulation
thread above sharpens this same attribution problem rather than
sidestepping it: a skill memory only compounds capability if the
system's judgment about which past attempts succeeded is itself
trustworthy, which routes straight back to the evaluation question of §7
and means accumulation is not a way around the harness-versus-model
question so much as a second place it resurfaces. Whether the harness
handling all of this was itself trained to handle long horizons, or is
purely inference-time scaffolding bolted onto a model trained for
something else, is the question §6 takes up.

\section{6. Training for Long
Horizons}\label{training-for-long-horizons}

The previous two sections describe scaffolding wrapped around a trained
model. This section asks what happens when the model itself is trained
with long horizons in mind, and its organizing observation is simple: as
horizon grows, outcome-only supervision --- one reward at the very end
of a long trajectory --- grows sparser and noisier per step, which is
exactly why process-level supervision and credit assignment have become
the field's central technical problem rather than a footnote to
reinforcement learning as usual (\texttt{rl} \(n=130\),
\texttt{supervision} \(n=37\) of the \texttt{training} category).

\textbf{Credit assignment under sparse, delayed reward.} When a single
trajectory reward must be distributed back across dozens of intermediate
actions, naive uniform credit assignment systematically mis-attributes
success and failure to steps that did not cause them. Segment Policy
Optimization addresses this directly by assigning credit at the level of
trajectory segments rather than whole trajectories or individual tokens,
on the premise that neither granularity extreme matches where causal
responsibility actually lives \citep{guo2025segment}; a graph-based
approach goes further, arguing that attribution restricted to a single
trajectory's own linear order --- even at segment granularity --- misses
credit relationships that only appear across a graph of related
trajectories \citep{cheng2026beyond}. Selective eligibility traces
revisit a much older RL idea specifically to avoid the opposite failure
of uniform credit assignment: spreading credit equally over every step
of a long trajectory dilutes the signal precisely where it is most
informative \citep{mou2026beyond}. Some of this thread ports
credit-assignment machinery into domains distant from its origin ---
multi-granularity intertemporal credit assignment for long-horizon
emotional-support dialogue, combining immediate and delayed credit from
a shared potential function over dialogue state \citep{zhang2026mica},
fair credit assignment for memory-augmented agents where the credit a
memory-write step deserves is entangled with whichever later step
actually uses that memory \citep{yan2026memory} --- evidence that credit
assignment has become a general-purpose long-horizon primitive rather
than a technique specific to any one task family. A comprehensive
empirical recipe for reinforcement learning on long-horizon tool-using
agents decomposes the agentic-RL design space along five axes --- reward
shaping, model scale, data composition, algorithm choice, and
environmental stability --- and is explicit that which choice is best is
itself scale-dependent (smaller models benefit from staged rewards and
extra exploration; larger models converge faster with simpler dense
rewards), meaning there is no single best recipe outcome-only RL alone
would reveal \citep{xixi2026demystifying}.

\textbf{Process reward models and generative supervision.} Where credit
assignment asks \emph{how to attribute} a sparse outcome signal, process
reward models ask whether a \emph{denser} signal can be trained directly
from step-level supervision instead of derived post hoc. The reference
result for this direction is Lightman et al.'s comparison of process-
against outcome-supervised reward models on the MATH dataset, which
reported that supervising each reasoning step significantly outperformed
supervising only the final answer --- but required 800,000 step-level
human feedback labels (released as PRM800K) to train the better reward
model \citep{lightman2023verify}. That pairing is the inheritance this
subsection works against: dense supervision demonstrably helps, at an
annotation cost no agentic setting can pay per task. Nearly everything
below is an attempt to obtain step-level density without step-level
human labels. Entropy-regularized process reward modeling directly
targets a distinct failure mode of naively trained PRMs ---
overconfident, poorly calibrated step scores --- by regularizing against
it during training \citep{zhang2024entropy}, and GroundedPRM grounds
step-level rewards in explicit tree-guided search plus a fidelity check,
precisely because ungrounded PRM training tends to reward
plausible-looking steps rather than steps that are actually correct
\citep{zhang2025groundedprm}. One striking observation ties this
subsection back to §6's opening claim about outcome supervision: GRPO
--- an algorithm introduced and typically described purely in
outcome-reward terms --- is shown to implicitly perform
process-reward-like credit attribution as a side effect of its
group-relative normalization, suggesting the boundary between
``outcome'' and ``process'' supervision is less an architectural choice
than a question of where in a training pipeline step-level credit ends
up being computed \citep{sullivan2025grpo}. Reward density is also being
pushed toward the long-horizon, agent-specific setting rather than
borrowed unmodified from short-horizon math/code reasoning: Think-RM
extends long-horizon reasoning into the generative-reward-model itself,
so that the judge doing the rewarding is trained with the same
long-horizon considerations as the policy it supervises
\citep{hong2025think}, AgentPRM builds step-wise promise-and-progress
signals specific to agent trajectories rather than reasoning chains
\citep{xi2025agentprm}, and SWE-TRACE combines rubric-based process
rewards with heuristics specifically to make long-horizon
software-engineering-agent training tractable \citep{han2026swe}. A
separate cautionary finding shows implicit process signal is not always
benign even when it exists: one paper documents agentic RL for search
actively misaligning a model's instruction-following behavior as an
unintended side effect of the reward it optimizes
\citep{yang2025agentic} --- a different mechanism from GRPO's
implicit-PRM structure above, but a reminder that reward shaping in
agentic RL has effects beyond the task metric it targets, a finding this
section flags rather than resolves, since it depends heavily on the
specific reward and environment involved.

\textbf{Boundary with the self-improvement literature.} Much of the
machinery in this section --- reward models, RL fine-tuning on
model-generated trajectories, iterative policy improvement --- overlaps
mechanically with the recursive-self-improvement literature surveyed in
a companion paper \citep{chen2026rsi}. The line we draw is one of
purpose, not mechanism: that survey treats these loops as an end in
themselves --- a system that trains on its own outputs, evaluates its
own quality signal, or improves its own research process, with the
interesting question being \emph{whether the loop closes}. Here, the
same techniques are training a policy to act well over more steps, with
a fixed external task-completion criterion the training never gets to
redefine --- the interesting question is \emph{how far one trained
policy can be pushed on longer tasks}, not whether the training loop
itself is self-referential. A single-agent RL run that uses
self-generated rollouts to train a longer-horizon tool-using policy
sits, by this criterion, inside this section; a system that additionally
modifies its own reward model, training procedure, or harness as part of
the same loop crosses into the companion survey's territory. Papers that
live on this boundary --- e.g., Q-Evolve, which unifies automatic
process-reward labeling with policy learning in a self-evolving,
in-distribution RL loop \citep{zhang2026selfa} --- are treated here
strictly as instances of longer-horizon policy training, leaving the
self-referential-loop question to the companion paper.

\textbf{Assessment.} The \texttt{rl}-to-\texttt{supervision} ratio (130
to 37) suggests the field has invested more effort in \emph{assigning}
credit within a fixed reward signal than in \emph{redesigning} the
reward signal's granularity outright --- the cheaper intervention, and
consistent with credit assignment's portability across the very
different domains cited above. But the GRPO finding above complicates a
clean read of that ratio: if outcome-level algorithms are already doing
implicit process-level work, the two subcategories are less separate
research programs than two vantage points on the same underlying problem
--- training a policy that gets useful signal from every step of a long
trajectory, not only its end. Whether that training actually produces
agents that are more reliable over long horizons, and how we would know,
is the measurement problem taken up next.

\section{7. Evaluation \& Measurement}\label{evaluation-measurement}

Every claim in §§3--6 is only as credible as the benchmark it was
measured against, and this section's organizing observation is that
agentic benchmarks have moved from single-step QA toward hours-long,
real-environment tasks faster than the methodology for validating them
has matured --- so that a large and increasingly self-aware slice of
this literature is now about whether the benchmarks measure what they
claim to (\texttt{evaluation}, entirely tagged \texttt{benchmark} in our
taxonomy, \(n=114\), plus the SWE-bench/GUI benchmark ecosystem that
recurs throughout \texttt{execution}). All benchmark numbers below are
dated at first mention, since public leaderboards shift on a timescale
of months.

\textbf{A case study in benchmark evolution: SWE-bench.} SWE-bench
\citep{jimenez2023swebench} --- resolve real GitHub issues given the
full repository as context --- has become the closest thing long-horizon
agent evaluation has to a common currency, and its evolution is a
microcosm of this section's argument. The original benchmark and its
immediate successors extended \emph{coverage}: SWE-bench-java
\citep{zan2024swe} and Multi-SWE-bench \citep{zan2025multi} add
languages beyond Python, SWE-bench Multimodal asks whether solutions
generalize to visual software domains the text-only original cannot
capture \citep{yang2024swe}, and SWE-Bench Pro raises the difficulty to
enterprise-scale, long-horizon problems explicitly because the
original's scope undersells the length of task real software engineering
requires \citep{deng2025swe}. A second, larger wave interrogates
\emph{validity} rather than coverage. SWE-bench+ manually inspects
successful patches and finds 32.67\% involve solution leakage (the fix
was already present in the issue report or comments) and a further
31.08\% pass only because the test suite is too weak to verify
correctness --- filtering both out drops one leaderboard-topping
system's resolution rate from 12.47\% to 3.97\% \citep{aleithan2024swe};
``Are `Solved Issues' in SWE-bench Really Solved Correctly?''
corroborates the second half of that finding independently, showing
passing patches diverge behaviorally from the human-written ground truth
in nearly 30\% of cases even when tests pass \citep{wang2025solved};
SWE-MERA responds by building a dynamic, continuously-updated benchmark
instead of a frozen one, arguing that the contamination and weak-test
problems SWE-bench+ documents are inherent to any benchmark that stops
collecting new issues after release \citep{adamenko2025swe}; and ``The
SWE-Bench Illusion'' shows state-of-the-art LLMs sometimes succeed by
recalling memorized repository content rather than reasoning about the
issue in front of them \citep{liang2025swe} --- a concern sharpened
further by a study asking directly whether SWE-Bench Verified tests
agent ability or model memory, and finding evidence for the latter in
frequently-benchmarked models \citep{prathifkumar2025swe}. A third
strand asks whether the leaderboards built on top of the benchmark are
themselves trustworthy: dissecting SWE-Bench's public leaderboards finds
systematic differences in how submitters profile and report results that
complicate cross-system comparison \citep{martinez2025dissecting},
corroborated by a follow-up comprehensive study of the same two
leaderboards \citep{martinez2026whats}. Most recently, a
trajectory-level diagnostic (TRAJEVAL) shows that even where capable
models localize the right code, they still fail after reaching it ---
``coherence collapse'' --- meaning Pass@1 alone actively misdiagnoses
\emph{why} the remaining 30--35\% of issues go unsolved
\citep{kim2026coherence}. Best-practice guidance distilled from across
this benchmark family generalizes the lesson beyond SWE-bench: across a
wider set of agentic benchmarks, task-setup and reward-design flaws
(SWE-bench Verified's insufficient test cases are exhibit one; a
separate example finds \(\tau\)-bench \citep{yao2024taubench} --- a
benchmark for tool-agent-user interaction under domain policies, notable
for scoring reliability across repeated trials of the same task rather
than single-shot success --- counts empty responses as successes) can
over- or under-estimate reported performance by up to 100\% in relative
terms, and a resulting checklist reduced measured overestimation on one
complex benchmark by 33\% when applied \citep{zhu2025establishing}.

\textbf{GUI and computer-use benchmarks.} A parallel ecosystem evaluates
agents that operate graphical interfaces rather than repositories.
WebArena \citep{zhou2023webarena} and OSWorld \citep{xie2024osworld}
established the realistic-environment paradigm for web and full-desktop
tasks respectively; AgentBench \citep{liu2023agentbench} and GAIA
\citep{mialon2023gaia} cover a wider span of agentic environments and
general-assistant tasks. The corpus's own supplement here is dense with
reliability findings that echo the SWE-bench pattern in a different
embodiment: OS-Harm evaluates the safety, not just the task success, of
computer-use agents specifically because prior work ``largely
overlooked'' harm potential while chasing capability
\citep{kuntz2025os}; AndroidControl-Curated finds that even top-scoring
GUI agents plateau around 60\% on AndroidControl not because the task is
unsolvable but because benchmark noise --- ambiguities and factual
errors in the benchmark itself --- caps the achievable score below what
the underlying capability would otherwise allow
\citep{leung2025androidcontrol}; and a direct study ``On the Reliability
of Computer Use Agents'' asks the sharpest version of this question: if
an agent succeeds at a task once, what stops it from succeeding again on
an unchanged retry, and attributes the answer to three factors:
stochasticity during execution, ambiguity in how the task itself is
specified (room for divergent-but-valid interpretations), and run-to-run
variability in the agent's own behavior --- reliability, in other words,
is not purely a property of the model, but of execution, task
specification, and agent behavior jointly
\citep{gonzalezpumariega2026reliability}. Bilingual and cross-platform
coverage is a growing concern rather than an afterthought: macOSWorld
extends the realistic-environment paradigm to a major OS the original
benchmarks omit \citep{yang2025macosworld}, and WindowsWorld argues that
single-application benchmarks like OSWorld understate the difficulty of
realistic, cross-application professional workflows
\citep{li2026windowsworld}.

\textbf{Measuring horizon directly: time-horizon metrics.} Rather than
asking whether an agent solves a fixed task, a distinct line of work
asks how \emph{long} a task an agent can complete at a given reliability
level, measured in human-time-to-complete --- an attempt to put a single
number on the ``how long a horizon can this system carry'' question this
whole survey is organized around. The most visible instance of this
measurement approach reports that the task length frontier models can
complete at a fixed success rate has grown at a measured, roughly
exponential rate over recent model generations \citep{kwa2025metr}. We
flag this explicitly as \emph{measured and reported}, not as a law:
doubling-type extrapolations of this kind are exactly the sort of
overclaim this survey's language calibration (§2.2) commits to avoiding,
and the metric's own assumptions --- what counts as ``the same'' task
across models of different eras, and whether human-time-to-complete on a
benchmark task translates to human-time-to-complete on the messier tasks
§1 opens with --- are contested rather than settled. We treat this
metric, and the critical literature interrogating agentic-benchmark
protocol validity more broadly \citep{zhu2025establishing}, as a
first-class thread rather than a premise: a benchmark from the SWE-bench
and GUI ecosystems above showing a capability plateau is only
informative to the extent the benchmark's own construction is
trustworthy, which is precisely what §7's second and third paragraphs
show cannot be assumed by default.

\begin{figure}
\centering
\includegraphics[width=0.85\linewidth,height=\textheight,keepaspectratio,alt={Figure 5: Human time-to-complete one task instance across the small set of benchmarks that report it directly, log scale. Dots are point estimates; bars are reported or illustrative ranges (see caption sources). Sourced entirely from primary papers -- artifacts/benchmark\_durations.csv.}]{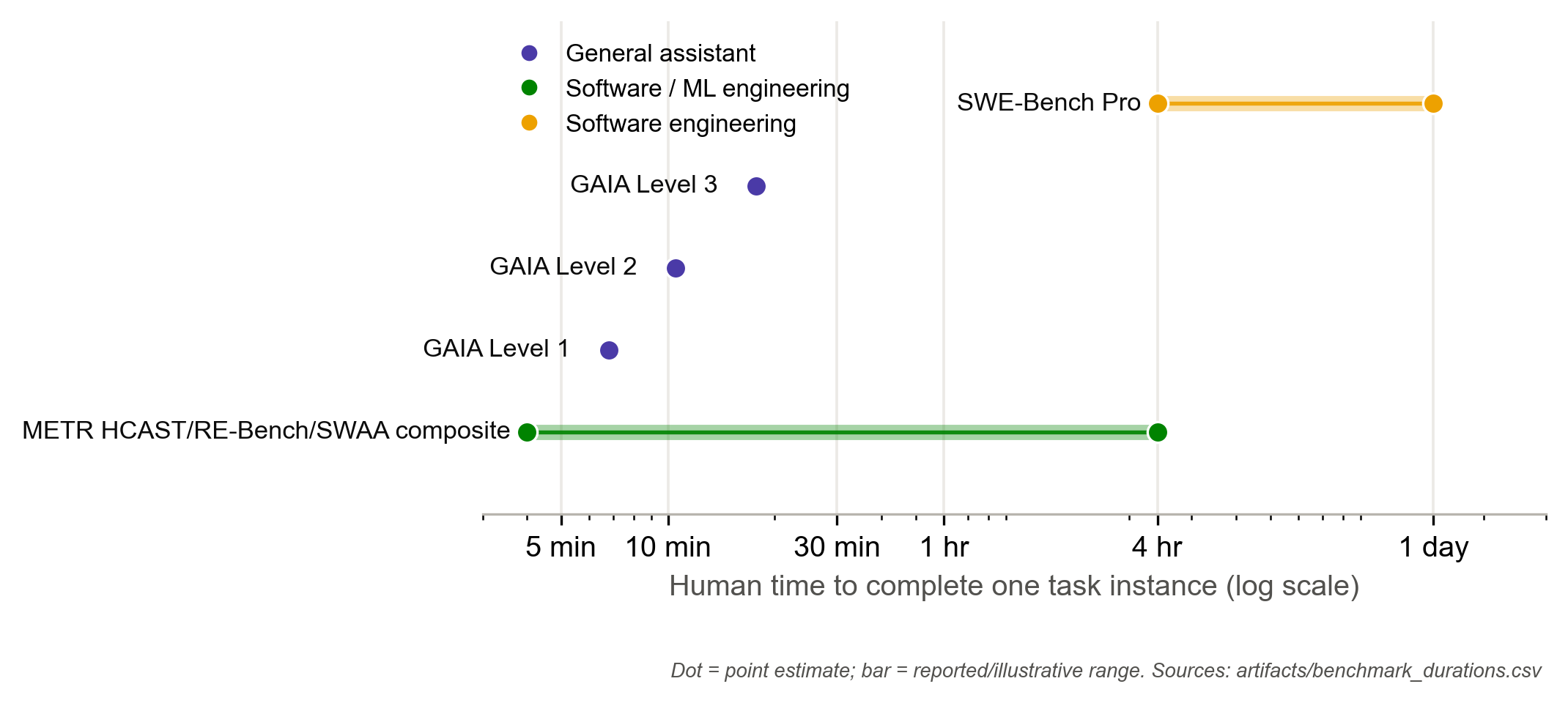}
\caption*{Figure 5: Human time-to-complete one task instance across the
small set of benchmarks that report it directly, log scale. Dots are
point estimates; bars are reported or illustrative ranges (see caption
sources). Sourced entirely from primary papers --
artifacts/benchmark\_durations.csv.}
\end{figure}

Figure 5 makes a point about the field's measurement maturity by its own
sparseness: of the roughly fifteen well-known agentic benchmarks we
checked against their own papers, only GAIA reports a clean,
level-by-level human-time-to-complete statistic; METR's own composite
task suite states the band over which model success degrades rather than
the suite's literal range; and SWE-Bench Pro states only a qualitative
``hours to days.'' SWE-bench itself, WebArena, and OSWorld --- three of
the most widely used agentic benchmarks in this literature --- report no
human-time baseline at all in their own papers. A field that wants
METR-style time-horizon measurement to generalize across benchmarks
needs more benchmarks to report the human-time statistic that
measurement depends on, not just more benchmarks.

\textbf{Evaluating what persists across tasks.} Every benchmark
discussed so far scores an agent on independent task instances, which by
construction cannot see whether an agent got better at the \emph{next}
task for having done the previous one. A small but distinct thread
builds benchmarks for exactly that. SWE-Bench-CL reorganizes SWE-bench
Verified's issues into chronologically ordered per-repository sequences,
so that experience accumulation, knowledge transfer across tasks, and
resistance to catastrophic forgetting become directly measurable rather
than incidental \citep{joshi2025swe}. The ELL framework pairs a
lifelong-learning agent architecture with a benchmark built on the same
premise --- that the interesting question for open-ended agents is
continuous growth through interaction, not performance on a static task
set \citep{cai2025building}. EvoAgentBench sharpens the target further,
arguing that what should transfer is \emph{procedural} --- reusable
searching, debugging, and verification routines extracted from execution
traces --- and that neither single-episode agent benchmarks nor
information-retention memory benchmarks isolate it
\citep{gao2026evoagentbench}. This thread is small (fewer than a dozen
papers in our corpus), which is itself notable: the systems literature
of §§3-5 is full of agents that claim to improve across sessions, but
the evaluation literature has only recently begun building instruments
that could confirm or refute those claims independently.

\textbf{Assessment.} The evaluation literature's shape mirrors the
training literature's from §6: just as outcome-only reward grows
uninformative as horizon grows, outcome-only benchmark scores (did the
patch pass the tests? did the agent reach the goal screen?) grow less
informative as the tasks being measured get longer and the ways to
superficially satisfy a check without solving the underlying problem
multiply. The field's response --- data purification, leakage audits,
trajectory-level diagnostics, reliability-under-retry studies, direct
human-time-based horizon measurement --- is evaluation's version of the
move from outcome to process supervision in §6, and the two are mutually
entangled in a way this section can describe but not resolve: better
benchmarks are needed to tell whether long-horizon training methods
work, and long-horizon training methods are increasingly what benchmark
designers use to probe where a benchmark's own construction breaks.
Whether that entanglement (§9 examines what it does and does not license
concluding), and the degradation patterns both this section and §6 keep
surfacing, admit any general theory is the question §8 takes up.

\section{8. Foundations, Limits \&
Safety}\label{foundations-limits-safety}

The preceding sections describe engineering responses to a common
pressure: as horizon grows, small per-step error rates compound, and
unattended systems accumulate risk that no single step reveals. This
section asks how far the field has gotten toward a general account of
that compounding, and toward the oversight problem it creates for agents
that run for a long time with no human watching every step. It is this
survey's smallest technical category by design (\(n=103\), after a
targeted supplement specifically because the seed harvest badly
under-covered it, §2.2), and its content is disclosed as more
theoretical, diagnostic, and safety-oriented than empirically settled.

\textbf{A simple model of compounding error, and where it breaks.} The
textbook picture of long-horizon degradation is straightforward: if
per-step error is \(\epsilon\) and errors compound independently,
success probability falls off exponentially in the number of steps --- a
model implicit in the ``why should a 100-step task be so much harder
than a 10-step task'' intuition motivating this whole survey. What the
corpus's diagnostic literature shows repeatedly is that real degradation
curves deviate from this simple picture in informative ways, on both
sides. Some deviations are worse than independent compounding predicts:
``Strained Coherence'' identifies a measurable pre-failure signal in
coding-agent execution trajectories --- trajectories that are about to
fail show detectable strain before the failure itself manifests, meaning
the error process is not memoryless the way the simple model assumes
\citep{pandya2026strained}, and ``Coherence Collapse'' (§7) shows
failures cluster at a specific stage of a trajectory rather than
distributing uniformly across steps \citep{kim2026coherence}. Other
deviations complicate the picture differently: Vending-Bench evaluates
long-term coherence directly (can an autonomous agent run a simulated
vending-machine business over an extended horizon --- runs exceeding 20M
tokens --- without losing track of its own state) and finds high
variance rather than smooth decay: capable models turn a profit in most
runs, but every model has some runs that derail into a ``meltdown'' loop
from which they rarely recover; notably, derailment shows no clear
correlation with the context window filling up, suggesting these
breakdowns are not simply a memory-capacity story
\citep{backlund2025vending}. Together these findings support a modest
but real conclusion: independent per-step error compounding is a useful
null model, not an empirically confirmed law, but the alternative it
points toward is not gentler degradation --- it is bimodal outcomes
(fine, or catastrophically derailed) whose trigger the field cannot yet
reliably predict.

\textbf{Goal drift and misalignment over long horizons.} A related but
distinct failure mode is not that an agent's actions become less
correct, but that its objective silently shifts. A technical report
evaluating goal drift in language model agents documents this directly
as a distinct phenomenon from accuracy decay \citep{arike2025technical},
and ``Governance Decay'' shows a specific, insidious mechanism for it:
context compaction --- the same compression §4 already treats as an
unresolved persistence-fidelity trade-off, not a solved one, even before
its safety implications are considered --- can silently erase the safety
constraints an agent was given at the start of a long trajectory,
precisely because a compression policy optimized for task-relevant
information has no reason to preserve constraints that never come up
again until they are violated \citep{chen2026governance}. Multi-agent
settings compound this further: ``The Coming Crisis of Multi-Agent
Misalignment'' argues that AI alignment work has not kept pace with the
dynamic, social nature of misalignment that emerges specifically from
\emph{interaction} between agents rather than from any single agent's
own objective \citep{carichon2025coming}. A formal treatment gives that
interaction-driven concern a specific mechanism: under a Bayesian model
of automated multi-agent workflows with weak per-workflow evidence, each
agent's generic pretraining prior over its own behavior dominates the
comparatively weak task-specific signal in its workflow prompt --- a
failure the paper names ``posterior collapse'' by analogy to the same
term in variational inference --- with the practical consequence that
distinct agents assigned distinct roles converge on nearly identical
actions, disregarding the role distinctions the workflow was designed
around; the paper argues this can be corrected only by injecting
context-specific evidence back into each agent's belief update, not by
better utility design alone \citep{ye2026sober}. Deception is a further
specific concern this literature has started to measure directly rather
than only theorize about: ``Cheap Talk, Empty Promise'' documents
frontier LLMs breaking public commitments for self-interest in
controlled settings \citep{shi2026cheap}, and a werewolf-game study
operationalizes deception and falsehood detection in a controlled
multi-agent environment specifically because naturalistic settings make
ground truth about an agent's beliefs hard to establish
\citep{mrinal2025wolf}.

\textbf{Oversight for long-running, unattended agents.} If an agent runs
unattended for a long horizon, the standard human-in-the-loop safety
pattern --- review every action before it executes --- is exactly what
``long-running'' and ``unattended'' preclude, which is why this
sub-thread treats oversight design as a first-order engineering problem
rather than a policy afterthought. Tiered Agentic Oversight proposes a
hierarchical multi-agent oversight structure modeled on clinical
hierarchy (nurse, physician, specialist) for safety-critical healthcare
settings, routing tasks to a tier by complexity; the lower tiers turn
out to be indispensable rather than redundant, absorbing up to 24\% of
individual agent errors before they can compound into a patient-facing
mistake, and removing them causes the largest safety degradation of any
ablation tested \citep{kim2025tiered}. Ensemble Monitoring for AI
Control finds that \emph{diversity} among monitors, not the number of
monitors or the compute spent on them, drives detection gains --- a
diverse three-monitor ensemble beats a homogeneous one built from three
copies of the same monitor by 2.4x, even at equal compute --- an
oversight-specific echo of the credit-assignment lesson from §6 that how
a signal is structured matters as much as how much of it there is
\citep{koran2026ensemble}. ``Managed Autonomy at Runtime'' proposes a
gear-based framework that adjusts an agent's operating autonomy level
dynamically rather than fixing it in advance, explicitly for single- and
multi-agent cyber-operations settings where the appropriate oversight
level is itself a function of context that changes during a long-running
task \citep{ramaswamy2026managed}. And a cautionary finding on the
limits of any single mitigation, this time upstream at the training-data
layer rather than at runtime: fine-tuning a model on synthetic agentic
trajectories that include adversarial actions measurably increases
misaligned behavior, and --- the ``phantom transfer'' finding --- this
increase survives \emph{removing every adversarial action from the
training trajectories before fine-tuning}, meaning the disposition
toward misalignment was encoded diffusely across the whole trajectory
rather than localized to the harmful steps a filter could catch
\citep{dixit2026filtering} --- direct evidence that safety interventions
aimed only at filtering visible harmful actions, whether at training
time or at runtime, are an incomplete solution to a problem that
originates upstream, in what the generating process teaches a model
about its own disposition.

Everything above is single-episode diagnosis: a run derails, a
constraint erodes, an oversight mechanism catches or misses one failure.
A separate and much smaller question is whether the capability to
\emph{avoid} these failures can itself accumulate across episodes, the
way §3, §5, §6, and §7 each found a minority thread doing for their own
topic. EvoAgentBench takes this on directly for foundational
measurement: it argues that what should transfer across a self-evolving
agent's episodes is not information but \emph{procedure} --- reusable
searching, debugging, and verification routines extracted from execution
traces --- and that this specific form of transfer is invisible to both
single-episode agent benchmarks and retention-focused memory benchmarks,
which is exactly why a dedicated instrument was needed
\citep{gao2026evoagentbench}. This is the one point in this section
where the cross-task-persistent locus of Axis 2 (§2.3) is addressed
directly rather than left for §9 to note as an absence; the fact that it
took a benchmark paper to raise it, rather than a theoretical account,
is itself consistent with this section's Assessment.

\textbf{Assessment.} No general, empirically validated theory of
long-horizon degradation currently unifies this section's findings; what
exists is a set of well-documented, partially contradictory phenomena
(bursty pre-failure signals, bimodal rather than smooth coherence loss,
drifting goals survivable by no single existing oversight mechanism)
that a future theory would need to explain simultaneously. This is this
survey's most explicit limitation, not a hedge: the category's small
size relative to the corpus, even after a targeted supplement, is itself
evidence that measurement and mitigation are ahead of theory here, and
the field's most useful near-term contribution may be sharper
diagnostics (§7's benchmarks, this section's pre-failure signals,
EvoAgentBench's procedural-transfer measurement) rather than a unifying
account of why long horizons are hard. We return to this gap, and to
what the corpus's own shape says about the field's priorities, in the
Discussion.

\section{9. Discussion}\label{discussion}

The body sections each ended with an Assessment specific to their own
literature. Four threads recur across multiple sections and are worth
naming explicitly, followed by an interpretive look at what the corpus's
own shape suggests about where the field is actually spending its effort
--- with the sampling caveats of §2.2 stated again first, since every
observation below is a hypothesis about the literature's composition,
not a finding about the underlying research problem's difficulty.

\textbf{Shared evaluator assumptions across training and evaluation.} §6
and §7 converge on the same structural problem from opposite ends.
Training methods for long horizons need a reliable way to tell whether a
trajectory is going well, and increasingly manufacture that signal via
process reward models and credit-assignment schemes that operate on
intermediate steps rather than final outcomes. Evaluation methods for
long horizons need exactly the same thing --- a reliable way to tell
whether a trajectory is going well, independent of whether it was
produced by a training method or an inference-time harness --- and
increasingly manufacture it via trajectory-level diagnostics and
process-aware benchmarks rather than terminal task success. The
implication worth flagging is that the tools used to validate whether
long-horizon training works (§7's process-aware benchmarks) and the
tools used to build long-horizon training in the first place (§6's
process reward models) rest on overlapping assumptions about what counts
as progress on a partial trajectory. We deliberately do not call this
circularity: nothing we observed shows a specific benchmark inheriting a
specific training signal, and independently constructed process signals
can and do cross-validate each other. The narrower and better-supported
concern is \textbf{correlated measurement bias} --- if the field's
shared intuitions about what ``good intermediate progress'' looks like
are systematically off in some respect, that error would be partly
invisible to a check that compares training methods against benchmarks
resting on the same intuitions. Establishing whether such a correlation
exists, and how strong it is, would require deliberately constructing
process signals from disjoint assumptions and comparing them --- an
experiment the corpus does not currently contain. We flag this as a
structural-analogy hypothesis rather than a finding, and it is worth
being explicit that it carries less evidentiary weight than the
harness-versus-model question below: nothing in the corpus demonstrates
this bias in a specific benchmark-training pair, whereas the
harness-versus-model question is backed by a concrete pattern in the
corpus itself (§5's orchestration-to-recovery ratio, §3's
decomposition-complexity finding). We keep both in this Discussion
because a structural risk worth naming does not have to be already
measured to be worth naming --- but the two should not be read as
equally supported.

\textbf{Harness vs.~model: where does long-horizon capability actually
live?} §3 and §5 both surface the same open attribution question from
different angles: how much of what makes a system complete a longer task
is the underlying model's capability, and how much is the harness
wrapped around it --- better context management, better orchestration,
better recovery logic --- with a fixed model underneath? §5's finding
that orchestration and recovery engineering vastly outnumber loop-level
model-capability papers in raw volume is suggestive but not dispositive;
it could mean the harness genuinely carries most of the load, or simply
that harness engineering is cheaper to iterate on than retraining a
model, so the corpus reflects research cost structure rather than where
capability actually lives. §3's finding that harnesses with more
elaborate decomposition are not uniformly better complicates the ``just
build a better harness'' reading further: past some point, additional
scaffolding measurably hurts rather than helps, meaning harness quality
itself has a capability ceiling that a fixed model cannot be scaffolded
past. Disentangling these two contributions --- the model's own
long-horizon competence versus the harness's compensation for its
absence --- is, in our reading, the single most consequential open
measurement problem this survey surfaces, because it bears directly on
whether long-horizon capability should be pursued by training better
models or building better harnesses, and the corpus does not yet contain
the controlled comparison that would answer it.

\textbf{Does long-horizon reliability scale?} §8 documents diagnostic
evidence on both sides of whether degradation over long horizons follows
any general, predictable curve --- bursty pre-failure signals and
drifting goals suggesting correlated, worse-than-independent error on
one side, high-variance bimodal outcomes that resist a single smooth
decay curve on the other --- without resolving to either. §7's measured,
reported (not proven) time-horizon growth trend \citep{kwa2025metr} is
the closest thing the field has to an aggregate answer, and it is
explicitly a capability trend across model generations, not a
reliability-scaling law within a fixed model or harness. Whether the
field will find a predictive theory here, or whether ``it depends on the
failure mode'' is the durable answer, remains open; we treat this as the
survey's most important flagged uncertainty rather than resolve it in
either direction.

\textbf{The trajectory is becoming the unit of analysis.} A fourth
pattern is visible only when the six sections are read together: across
every category, the object being stored, scored, attributed over,
audited, and debugged is shifting from the \emph{outcome} to the
\emph{execution trace}. In training, credit assignment operates over
trajectory segments and trajectory graphs rather than final rewards (§6)
--- SALT assigns step-level advantage by building a graph over the
trajectory itself \citep{jiazheng2025salt}. In evaluation, Pass@1 is
being displaced by trajectory-level diagnosis, and the
coherence-collapse finding was only visible \emph{because} the
diagnostic decomposed 16,758 stored trajectories into aligned stages
(§7) \citep{kim2026coherence}. In memory, the question has moved from
what to store to what a stored trace can later be held accountable for:
MemWeaver builds a memory framework explicitly for ``traceable''
long-horizon reasoning with evidence-grounded reuse
\citep{ye2026memweaver}, and Slipstream validates context compaction
against the trajectory it compacts, treating the compactor's unawareness
of what will later be needed as a structural validation gap
\citep{chen2026slipstream}. In safety, forensic trajectory signatures
make memory poisoning detectable after the fact (§8)
\citep{leong2026forensic}. And in the self-evolution literature that
cuts across §§6-8, EvoAgentBench argues the transferable unit is
\emph{procedural} --- reusable search, debug, and verification
procedures extracted from execution traces --- which existing
single-episode agent benchmarks and information-retention memory
benchmarks both fail to isolate \citep{gao2026evoagentbench}. This
convergence is worth naming because it has a practical consequence the
individual sections do not state: trajectory retention is quietly
becoming a precondition for progress on the other three threads above.
The harness-versus-model attribution question (thread two) is not
answerable from outcome scores alone, and the
correlated-measurement-bias concern (thread one) can only be tested by
comparing process signals against each other at the step level. The
reliability-scaling question (thread three) depends on it too, and in
the most direct way of the three: §8's own evidence that degradation is
bursty and bimodal rather than smooth --- the pre-failure signal in
Strained Coherence, the meltdown-or-profit split in Vending-Bench --- is
itself trajectory-level diagnosis, not an outcome statistic; a field
that only logged pass/fail per episode would not have been able to
observe either finding. A field that discards its execution traces
cannot run any of these three experiments --- which makes trajectory
logging, and the storage and privacy questions that come with it (§4,
§8), infrastructure rather than an implementation detail.

\begin{figure}
\centering
\includegraphics[width=0.95\linewidth,height=\textheight,keepaspectratio,alt={Figure 6: Growth timeline, seed corpus only (n=1,419). Top: quarterly paper counts per category, log scale. Bottom: category share of quarterly output. The partial final quarter (2026Q3, 76 papers) is omitted from the trend and annotated separately.}]{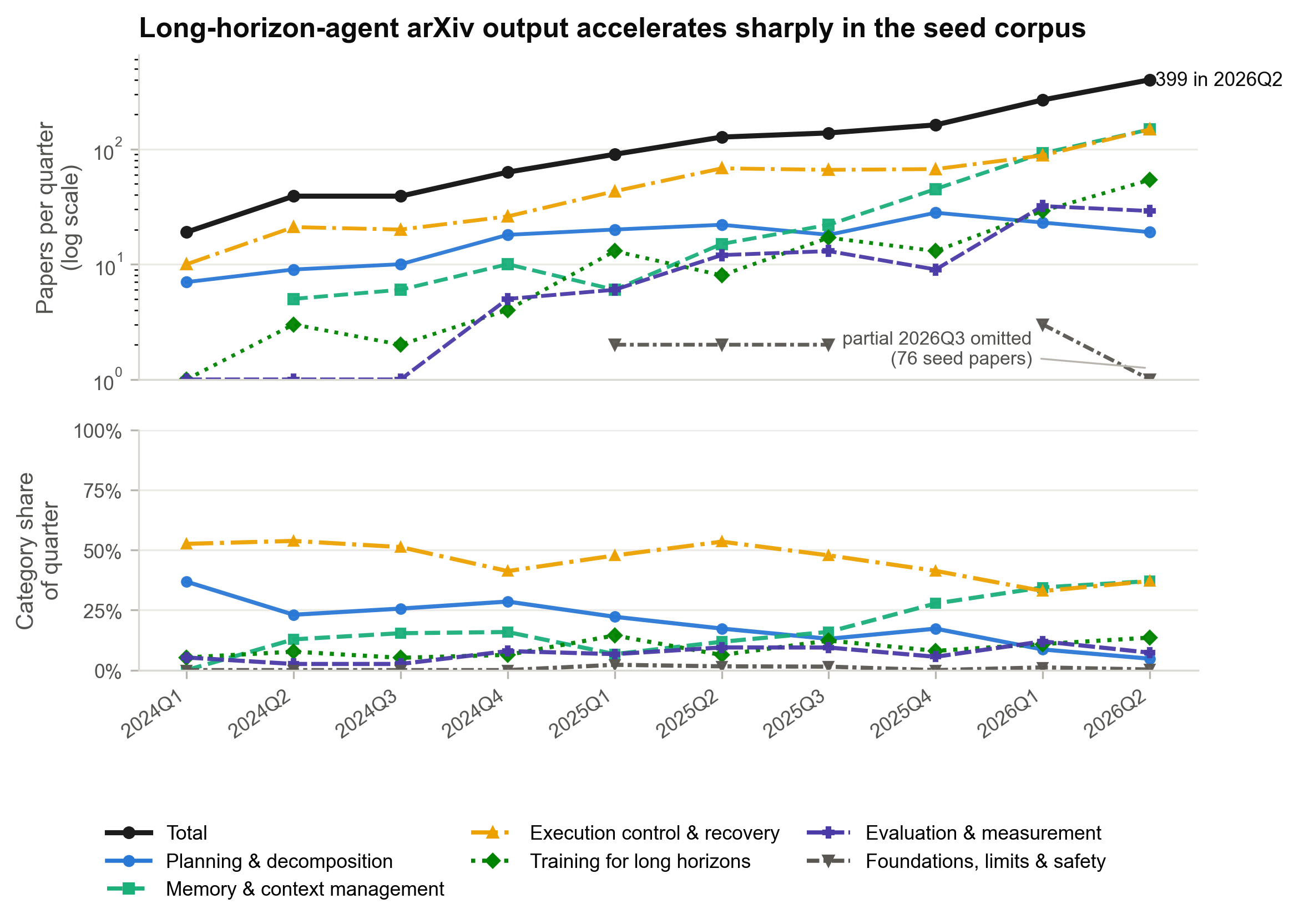}
\caption*{Figure 6: Growth timeline, seed corpus only (n=1,419). Top:
quarterly paper counts per category, log scale. Bottom: category share
of quarterly output. The partial final quarter (2026Q3, 76 papers) is
omitted from the trend and annotated separately.}
\end{figure}

\textbf{What the corpus's shape suggests (hypotheses, not findings).}
Figure 6's two panels show \texttt{execution} as the largest category by
share throughout 2024-2025 --- consistently the plurality leader,
fluctuating roughly 41-54\% quarter to quarter rather than holding a
flat majority --- and \texttt{memory}'s share rising to match or
slightly exceed it from 2026 Q1 onward, while \texttt{planning}'s share
has drifted downward over the same period. One explanation consistent
with this pattern: the field solved a version of ``make an agent take
more than one action'' (execution/orchestration) early and cheaply, and
only once agents were routinely attempting genuinely long trajectories
did the specific failure mode of losing track of earlier information
(memory) become salient enough to attract comparable research volume ---
a herd-dynamics-style account in which research attention follows
visible failure, not a claim that memory is intrinsically harder than
execution. Table 1's 2026-share column offers a second, complementary
hypothesis: \texttt{foundations} (77\% posted in 2026) and
\texttt{memory} (72\%) skew far more recent than \texttt{planning}
(27\%), consistent with a verifiability-gradient explanation ---
decomposition strategies are comparatively easy to demonstrate and
publish quickly, while diagnosing why a system fails over a long
horizon, or building memory architectures robust to interference at
scale, requires the kind of large-scale deployment experience that has
only become available recently. Both explanations are offered as
testable hypotheses, in the sense that they predict specific further
patterns (e.g., a verifiability-gradient account predicts foundations
papers should lag their own triggering capability demonstrations by a
roughly consistent delay), not as conclusions this corpus alone can
establish.

\section{10. Conclusion}\label{conclusion}

This survey organized 1,547 arXiv papers (2024-2026) around a single
question: as a task's horizon grows past what one context window or one
uninterrupted attempt can hold, what breaks first, and what compensates?
Planning trades early commitment for robustness as environments grow
less predictable (§3); memory trades fidelity for persistence across a
three-tier hierarchy of context, external storage, and weights, with no
tier escaping that trade-off (§4); execution control increasingly relies
on the surrounding harness --- orchestration and recovery --- more than
on the underlying model's own reliability, though the two remain hard to
disentangle (§5); training is moving from sparse outcome signals toward
denser, step-level credit as horizons lengthen, blurring the line
between ``outcome'' and ``process'' supervision (§6); evaluation is
undergoing the same shift, with a substantial and growing share of the
literature devoted to showing that existing benchmarks do not measure
what they claim to (§7); and the theory needed to unify these
observations into a predictive account of long-horizon degradation does
not yet exist, though the diagnostic and safety literature building
toward it is real and growing (§8). None of these six threads is
complete on its own, and the Discussion's cross-cutting observations ---
the risk of correlated measurement bias between training and evaluation
signals, the harness-versus-model attribution problem, whether
long-horizon reliability scales at all, and the field's convergence on
the execution trajectory as its shared unit of analysis --- are, in our
assessment, where the field's next real progress has to be made. We hope
the taxonomy and the corpus behind it (§ Data Availability) make that
progress easier to track.

\section{Data Availability}\label{data-availability}

The corpus (\texttt{corpus\_v2.csv}, 1,547 papers with taxonomy labels),
the classification and figure-generation scripts, and the full
bibliography (\texttt{references.bib}, auto-generated;
\texttt{anchors.bib}, hand-curated) will be made available at a public
repository accompanying this paper, to be linked in the
camera-ready/published version. Classification rules, the harvest query
threads, and the bleed-quantification procedure described in §2.2 are
released as executable scripts rather than described only in prose, so
that the corpus can be regenerated, audited, or extended.

\section{References}\label{references}

\bibliography{references}

\end{document}